\documentclass{article}

\usepackage{stfloats}
\usepackage{cuted}
\usepackage{caption}

\usepackage{multirow}
\usepackage{graphicx}
\usepackage{booktabs}
\usepackage{tabularx}
\usepackage{array}
\usepackage{pifont}
\usepackage{enumitem}

\usepackage[table]{xcolor}
\usepackage{pifont}
\usepackage{booktabs}
\usepackage{tabularx}
\usepackage{wrapfig}

\newcommand{\cmark}{\textcolor[HTML]{47D45A}{\ding{51}}}
\newcommand{\xmark}{\textcolor[HTML]{FF0000}{\ding{55}}}

\usepackage{textcomp}
\usepackage{siunitx}

\usepackage{listings}
\usepackage{xcolor}
\usepackage{textcomp}

\definecolor{codetext}{HTML}{2F3337}
\definecolor{codekeyword}{HTML}{FF8887}
\definecolor{codecommand}{HTML}{81BBE0}
\definecolor{codestring}{HTML}{6F9FBF}
\definecolor{codecomment}{HTML}{8A8F94}
\definecolor{codenumber}{HTML}{B0B4B8}

\lstdefinestyle{paperbash}{
    language=bash,
    basicstyle=\ttfamily\bfseries\fontsize{10}{12}\selectfont\color{codetext},
    keywordstyle=\bfseries\color{codekeyword},
    keywordstyle=[2]\bfseries\color{codecommand},
    commentstyle=\bfseries\color{codecomment},
    stringstyle=\bfseries\color{codestring},
    alsoletter={-},
    morekeywords=[2]{mmseqs,createdb,createindex,easy-search},
    backgroundcolor=\color{white},
    frame=none,
    xleftmargin=26pt,
    xrightmargin=0pt,
    numbers=left,
    numberstyle=\ttfamily\bfseries\fontsize{7}{9}\selectfont\color{codenumber},
    numbersep=10pt,
    stepnumber=1,
    numberblanklines=false,
    breaklines=true,
    breakatwhitespace=true,
    breakautoindent=true,
    columns=fullflexible,
    keepspaces=true,
    tabsize=4,
    showspaces=false,
    showtabs=false,
    showstringspaces=false,
    aboveskip=0.8em,
    belowskip=0.8em
}

\usepackage{microtype}
\usepackage{graphicx}
\usepackage{subcaption}
\usepackage{booktabs} 

\usepackage{hyperref}

\usepackage[preprint]{icml2027}

\usepackage{amsmath}
\usepackage{amssymb}
\usepackage{mathtools}
\usepackage{amsthm}

\usepackage{enumitem}

\usepackage[capitalize,noabbrev]{cleveref}

\theoremstyle{plain}

\theoremstyle{definition}

\theoremstyle{remark}

\usepackage[textsize=tiny]{todonotes}

\icmltitlerunning{Submission and Formatting Instructions for ICML 2027}

\begin{document}

\twocolumn[
  \icmltitle{CryoProt: A Protein Pretraining Framework with Cross-Box \\Interactions on Cryo-EM Density Maps}



  \icmlsetsymbol{equal}{*}

  \begin{icmlauthorlist}
    \icmlauthor{Dan Luo}{hnu,xtu}
    \icmlauthor{Xuan Lin}{xtu}
    \icmlauthor{Peng Zhou}{hnu}
    \icmlauthor{Junwen Zhu}{hnu}
    \icmlauthor{Tengfei Ma}{hnu}
    \icmlauthor{Xiangxiang Zeng}{hnu}
    \icmlauthor{Yiping Liu}{hnu}
  \end{icmlauthorlist}

  \icmlaffiliation{hnu}{College of Computer Science and Electronic 
Engineering, Hunan University, Changsha, China}

  \icmlaffiliation{xtu}{School of Computer Science, Xiangtan University, 
Xiangtan, China}

  \icmlcorrespondingauthor{Xuan Lin}{jack\_lin@xtu.edu.cn}
  \icmlcorrespondingauthor{Yiping Liu}{yiping0liu@gmail.com}

  \icmlkeywords{Cryo-EM Density Map, Protein representation, Pretraining}

  \vskip 0.3in
]



\printAffiliationsAndNotice{}  

\begin{abstract}
  Despite the growing availability of cryo-electron microscopy (cryo-EM) density maps, effectively leveraging them for protein representation remains challenging. First, current methods lack a general-purpose protein pretraining framework tailored for cryo-EM density maps, designed for protein-related property prediction. Second, existing approaches typically partition density maps into local box regions and model them independently, overlooking interactions across boxes which are essential for capturing global structural context in cryo-EM density map. To address these challenges, we propose \textbf{CryoProt}, a protein pretraining framework for cryo-EM density maps. Its MLA-based Map Encoder explicitly models cross-box dependencies in latent space, while multi-task pretraining learns transferable representations for downstream tasks such as protein flexibility prediction, which requires no density maps at inference. Experimental results demonstrate that CryoProt consistently outperforms existing state-of-the-art methods across multiple benchmarks, achieving up to 12\% improvement over the best-performing baselines, highlighting the importance of modeling cross-box interactions in cryo-EM data. The source code is publicly available at \url{https://anonymous.4open.science/r/CryoProt}.
\end{abstract}

\section{Introduction}

Proteins carry out a wide range of essential biological processes \citep{truong2023poet, madsen2025topological}, and effective protein representation is fundamental for understanding and predicting their functional properties \citep{duan2025boosting, detlefsen2022learning}. Existing protein representations primarily rely on either sequence information or structural information \citep{han2025copra}. The former leverages large-scale amino acid sequences to capture evolutionary and contextual information \citep{rao2021msa}, while the latter utilizes three-dimensional coordinates of residues to model spatial organization and geometric relationships \citep{yuan2025protein}. Both paradigms rely on abstracted representations that are either indirectly inferred or computationally predicted, which may fail to fully capture the underlying physical properties of biomolecules.

\begin{figure*}[t]
  \centering
  \includegraphics[width=1.0\linewidth]{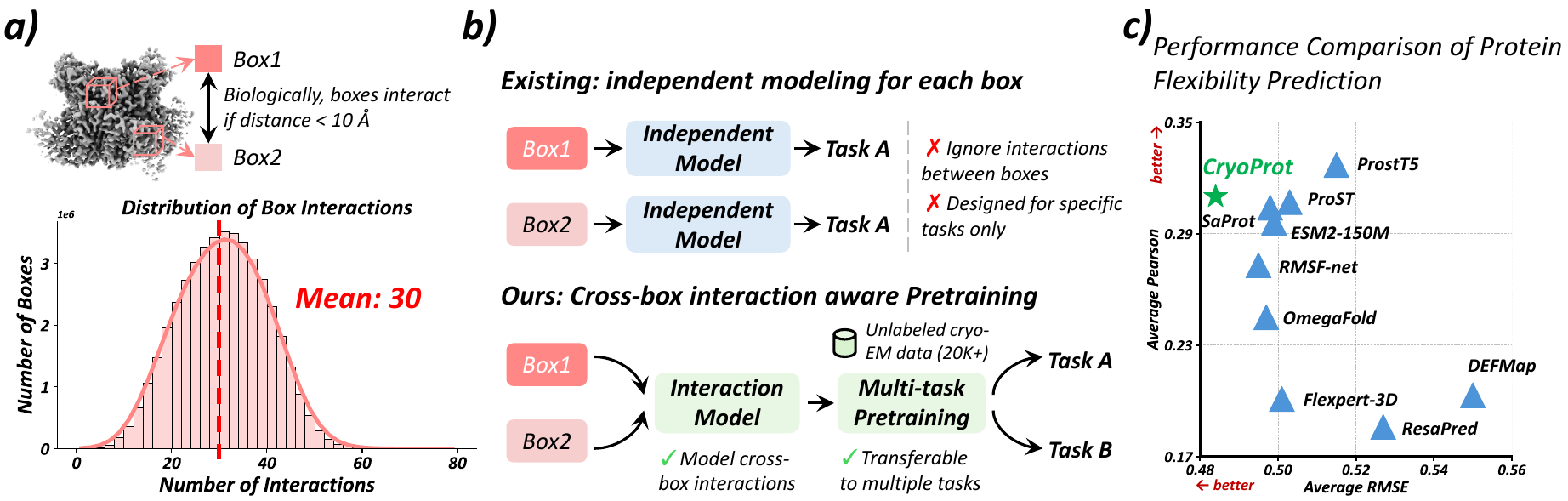}
  \caption{(a) The upper sub-figure illustrates how interactions between local box regions in a density map are determined. From a biological perspective, boxes are considered interacting if their distance is less than 10 Å. The lower sub-figure shows the distribution of the number of boxes under different interaction counts.
  (b) Existing methods model each box independently, whereas the proposed pre-trained CryoProt considers interactions across boxes.
  (c) Performance comparison between CryoProt and state-of-the-art methods on the protein flexibility prediction task.}
  \label{fig:our_work}
\end{figure*}

In contrast, cryo-electron microscopy (cryo-EM) offers a more direct and physically grounded modality. Recent advances in cryo-EM have led to a rapid increase in the availability of high-resolution density maps \citep{kimanius2024data, he2024accurate}, which approximate the electron scattering potential of biomolecules and provide a continuous three-dimensional view of their spatial organization \citep{raghu2025multiscale}. Unlike sequence or discretized structural representations, cryo-EM density maps preserve rich low-level structural signals, capturing both local environments and global conformational variations \citep{he2024accurate, sanchez2021deepemhancer, liu2025cryoalign2}, thereby enabling the learning of generalizable protein representations. However, existing approaches leveraging cryo-EM density maps still face two critical challenges that limit their effectiveness in modeling proteins.

\textit{Challenge 1: Lacking general-purpose protein pretraining models for cryo-EM density maps.} Existing methods leveraging cryo-EM maps are often tailored to specific downstream applications, such as protein
flexibility prediction \citep{matsumoto2021extraction, song2024accurate, vander2024atlas}. While these task-specific designs achieve promising performance within their respective domains, they lack a unified framework for learning transferable representations across diverse tasks. Consequently, the rich structural information in cryo-EM density maps is not fully utilized, and the resulting models often show limited generalization, making it difficult to develop scalable pretraining frameworks for cryo-EM data.

\textit{Challenge 2: Overlooking interactions across local boxes in density map.} As shown in the upper panel of Fig.~\ref{fig:our_work}(b), many existing methods partition cryo-EM density maps into local cubic boxes of predefined sizes for localized processing
\citep{chen2024protein, selvaraj2025cryoten, zhou2024cryofm}. While this design reduces computational cost, it overlooks interactions across boxes. In practice, protein structures are inherently interconnected, and spatial dependencies between regions are important for capturing structural and functional properties \citep{koren2023intramolecular, jumper2021highly}. As illustrated in Fig.~\ref{fig:our_work} (a), each box interacts with around 30 neighboring boxes on average. Neglecting such dependencies can lead to incomplete structural representations, as the model fails to capture global structural context, ultimately degrading downstream performance.

To address these challenges, we propose CryoProt, a novel general-purpose protein pretraining framework built directly upon cryo-EM density maps. CryoProt employs a Map Encoder based on multi-head latent attention (MLA) to effectively model density maps and capture complex interactions across local boxes in shared latent space. Furthermore, we design a multi-task pretraining strategy that enables the model to learn generalizable representations, which can be readily transferred to diverse downstream tasks. As illustrated in Fig.~\ref{fig:our_work} (c), protein flexibility prediction is shown as a representative downstream task, where CryoProt achieves state-of-the-art performance. Across a wide range of downstream benchmarks, CryoProt consistently outperforms existing methods, demonstrating its strong generalization capability. Our contributions are as follows:
\begin{itemize}[leftmargin=*]
    \item We propose CryoProt, the first general-purpose protein pretraining framework that explicitly models cross-box interactions in cryo-EM density maps, enabling the learning of transferable representations from large unlabeled cryo-EM data for diverse downstream tasks.
    
    \item We design an MLA-based Map Encoder for explicit cross-box interaction modeling, where box-level representations interact through a shared latent space. Furthermore, we develop a multi-task pretraining strategy to enhance representation learning and generalization.
    
    \item CryoProt achieves state-of-the-art performance across multiple benchmarks, with up to 12\% improvement over the best-performing baselines. Ablation studies and visualization analyses further validate the effectiveness of modeling cross-box interactions in cryo-EM data.
\end{itemize}

\section{Related Work}

\textbf{Protein Pretraining Models.}
Sequence-based pretrained models such as ESM-1b \citep{rives2021biological}, ESM-2 \citep{lin2022language} and ESM-3 \cite{hayes2025simulating} use Transformers to capture evolutionary information. ProtST \citep{xu2023protst} enhances protein representations by jointly learning from protein sequences and biomedical texts. Beyond sequence-based approaches, structure-aware models are explored. OmegaFold \citep{wu2022high} employs memory-efficient self-attention without multiple sequence alignments \citep{zielezinski2025ultrafast}, while SaProt \citep{su2024saprot} uses a structure-aware vocabulary and ProstT5 \citep{heinzinger2024bilingual} learns bidirectional translation between protein sequences and structures. Most methods rely on sequence or structure, with limited use of experimental properties such as cryo-EM density maps.

\textbf{Protein-related Property Prediction.} Numerous task-specific models have been proposed for predicting protein-related properties, including residue-level tasks (e.g., protein flexibility prediction \citep{luo2026dynamicdta} and identification of active sites) and protein-level tasks (e.g., protein--protein binding affinity prediction and mutation-induced binding free energy change $\Delta\Delta G$ prediction). For flexibility prediction, ResaPred \citep{wang2025resapred} employs residual networks with self-attention to model sequence features, while Flexpert-3D \citep{kouba2025learning} integrates sequence and structural information. For site identification, Deep-ProBind \citep{khan2025deep} combines transformer-based encoding with handcrafted features, whereas MMSite \citep{ouyang2024mmsite} and M$^3$Site \citep{ouyang2025m3site} adopt multimodal frameworks to incorporate structural and functional context. DeepProSite \citep{fang2023deepprosite} integrates predicted structures with pretrained protein language model representations for binding site prediction. For binding affinity prediction, Bind-ddG \citep{shan2022deep} adopts an attention-based geometric neural network, while GearBind \citep{cai2024pretrainable} and PPIgraphomer \citep{xie2025ppi} utilize geometric graph neural networks and graph Transformers, and Island \citep{abbasi2020island} relies on sequence features for regression. For $\Delta\Delta G$ prediction, PPIformer \citep{bushuiev2023learning} introduces an SE(3)-equivariant architecture, ProBASS \citep{gurusinghe2025probass} leverages pretrained protein language models, DGCddG \citep{jiang2023dgcddg} applies graph convolutional network to model mutation effects, and MFFN \citep{zhang2025multi} employs multi-scale feature fusion with attention mechanisms.

\textbf{Modeling Based on Cryo-EM Density Maps.}
Existing studies on cryo-EM density maps can be divided into two directions. The first focuses on map reconstruction and resolution enhancement, with methods such as CryoFM \citep{zhou2024cryofm}, and EModelX \citep{chen2024protein}, often partitioning maps into local box regions to reduce computational cost. The second explores modeling protein-related properties from cryo-EM data, which remains less studied. Methods such as DEFMap \citep{matsumoto2021extraction} extract residue-centered regions and apply 3D convolutional networks to predict flexibility, while RMSF-Net \citep{song2024accurate} incorporates simulated density maps for improved performance. However, these approaches are task-specific and lack generalization. To address this limitation, we propose CryoProt to leverage cryo-EM density maps for general protein property prediction across diverse downstream tasks.

\section{Method}
\label{method}

\subsection{Dataset Construction}
\label{dataset}

The pretraining dataset is constructed from the Electron Microscopy Data Bank (EMDB) \citep{lawson2016emdatabank}. Following prior studies \citep{matsumoto2021extraction, chen2024protein}, we curate 20,530 high-quality cryo-EM protein samples, forming one of the largest curated datasets of this type compared with previous studies \citep{matsumoto2021extraction, dhakal2024cryotransformer, zhang2025emol}. To evaluate generalizability, we consider four downstream tasks spanning residue- and protein-level prediction, namely protein flexibility prediction, active site identification, binding affinity prediction, and $\Delta\Delta G$ prediction. The corresponding datasets are RMSFNet \citep{song2024accurate}, ProTAD \citep{ouyang2024mmsite}, HER2 \citep{shanehsazzadeh2023unlocking}, and SKEMPI v2 \citep{jankauskaite2019skempi}, respectively. Exact protein overlap between pretraining and downstream datasets is excluded, and an additional MMseqs2-based\footnote{\url{https://github.com/soedinglab/MMseqs2}} analysis shows that sequence similarity between downstream proteins and the pretraining set is generally limited. All downstream experiments use five-fold cross-validation \citep{alquraishi2019proteinnet, vskrhak2025cryptobench}. The statistics of the pretraining and downstream datasets are summarized in Table~\ref{tab:datasets}. Detailed dataset construction and similarity analysis are provided in Appendix~\ref{appendix:dataset}.

\begin{table}[!h]
\centering
\caption{Summary of datasets in this study.}
\label{tab:datasets}
\small
\setlength{\tabcolsep}{1.5pt}
\renewcommand{\arraystretch}{1}
\begin{tabular}{c c c c c c}
\toprule
 & \multicolumn{1}{c}{\textbf{Pre-training}} & \multicolumn{4}{c}{\textbf{Downstream tasks}} \\
\cmidrule(lr){2-2} \cmidrule(lr){3-6}
\textbf{Dataset} & EMDB & RMSFNet & ProTAD & HER2 & SKEMPI v2 \\
\midrule
\textbf{Entries} & 20,530 & 335 & 6,318 & 422 & 7,208 \\
\textbf{Level}   & --     & Residue & Residue & Protein & Protein \\
\bottomrule
\end{tabular}
\end{table}

\subsection{CryoProt Framework}

\begin{figure*}[!t]
\centering
\includegraphics[width=0.9\linewidth]{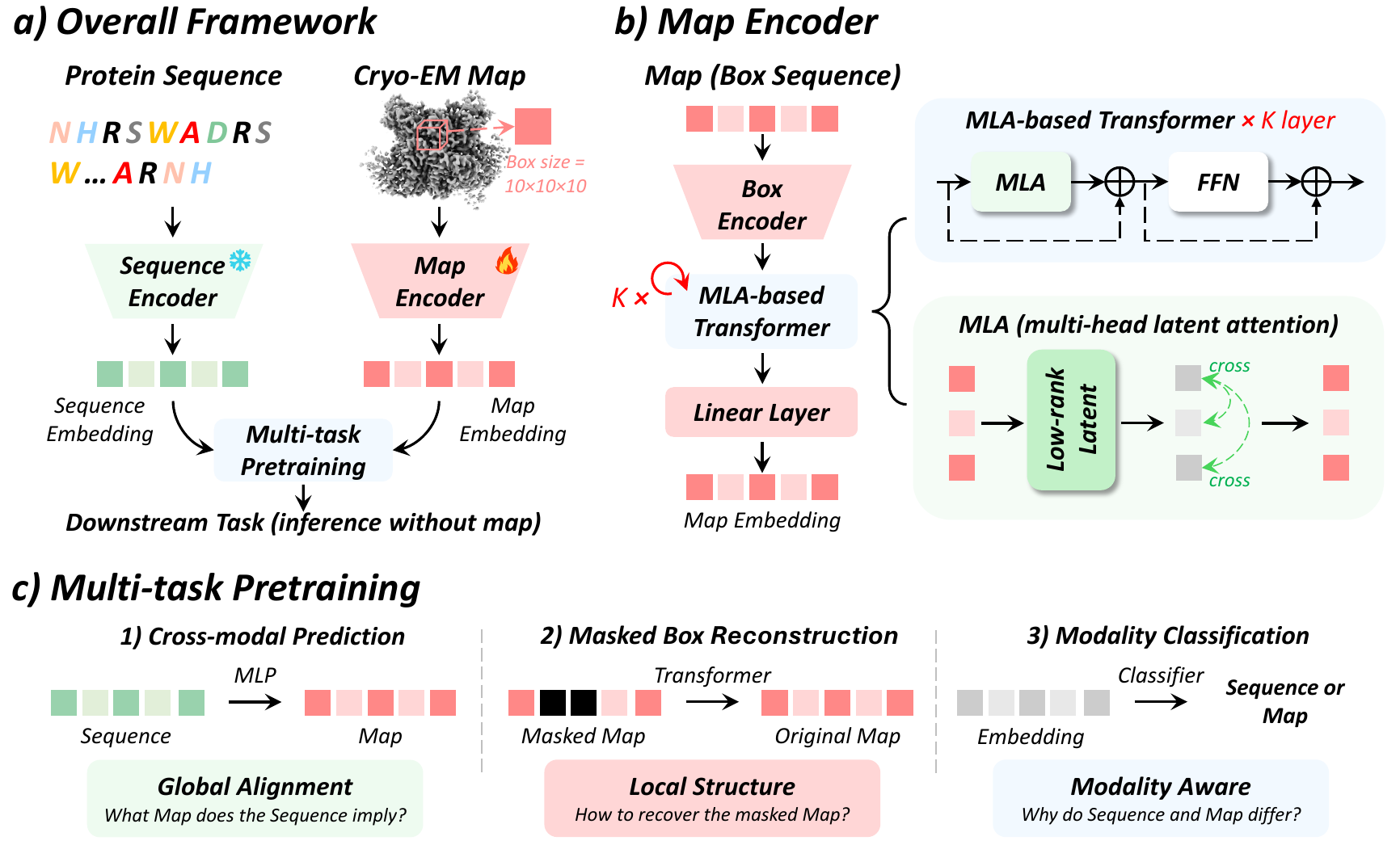}
\caption{(a) Overview of the CryoProt framework, which employs an MLA-based Map Encoder to capture cross-box interactions and incorporates multi-task pretraining to learn transferable protein representations. (b) The proposed Map Encoder. (c) The designed multi-task pretraining strategy.}
\label{fig:framework}
\end{figure*}

As illustrated in Fig.~\ref{fig:framework}, the pretraining protein model CryoProt consists of key components (e.g., a Sequence Encoder for sequence-level representations, a Box Encoder for local structural contexts, and a Map Encoder for global cross-box interactions). Details are described below.

\textbf{Sequence Encoder.}
Given a protein sequence $S = \{a_1, a_2, \dots, a_L\}$, we first tokenize it into amino acid tokens and encode it using a frozen ESM2-150M model~\citep{lin2022language}, yielding contextual residue-level representations $\mathbf{E}^{\text{seq}} \in \mathbb{R}^{L \times 640}$. To align feature dimensions with the cryo-EM Map Encoder, we apply a linear projection followed by a ReLU activation function
\begin{equation}
\tilde{\mathbf{E}}^{\text{seq}} = \mathrm{ReLU}\left(\mathbf{W}_s \, \mathbf{E}^{\text{seq}} + \mathbf{b}_s\right),
\end{equation}
where $\mathbf{W}_s$ and $\mathbf{b}_s$ are learnable parameters, with $\tilde{\mathbf{E}}^{\text{seq}} \in \mathbb{R}^{L \times 240}$. We set the maximum length to 1024. Sequences exceeding this threshold are excluded during preprocessing, while shorter ones are zero-padded and accompanied by an attention mask for valid token-level computation.

\textbf{Box Encoder.}
We first represent each cryo-EM density map as a set of residue-centered local sub-volumes. Given a map representation $N$, we partition it according to residue positions into a set of local boxes $N \rightarrow \{\text{box}_1, \text{box}_2, \dots, \text{box}_L\}$. Each $\text{box}_i \in \mathbb{R}^{10 \times 10 \times 10}$ encodes the local structural environment of residue $i$ and is subsequently reshaped into a vector $\mathbf{E}_i^{\text{box}} \in \mathbb{R}^{1000}$. To incorporate spatial information, we encode the 3D coordinates of each box within the cryo-EM density map $\mathbf{C}_i = (x_i, y_i, z_i)$ using sinusoidal positional encoding \citep{vaswani2017attention}
\begin{equation}
\begin{split}
\text{PE}(c_i, 2k) &= \sin\left(\frac{c_i}{10000^{2k/d}}\right), \\
\text{PE}(c_i, 2k+1) &= \cos\left(\frac{c_i}{10000^{2k/d}}\right),
\end{split}
\end{equation}
where $c_i \in \{x_i, y_i, z_i\}$ and $k$ indexes the embedding dimension. The structural and positional information are first fused, and then projected into a latent space via a learnable MLP
\begin{equation}
\tilde{\mathbf{E}}_i^{\text{box}} = \text{MLP}\left( \mathbf{E}_i^{\text{box}} + \text{PE}(\mathbf{C}_i) \right).
\end{equation}

\textbf{Map Encoder.}
After the box encoding stage, the cryo-EM density map is represented as a sequence of box embeddings $\mathbf{E}^{\text{map}} = \{\tilde{\mathbf{E}}_1^{\text{box}}, \tilde{\mathbf{E}}_2^{\text{box}}, \dots, \tilde{\mathbf{E}}_L^{\text{box}}\}$. To model global dependencies and cross-box interactions, we design a Transformer-based architecture equipped with Multi-Head Latent Attention (MLA) \citep{liu2024deepseek}. Although MLA retains the quadratic complexity of conventional attention, it projects box embeddings into a lower-dimensional latent space to obtain compressed representations $\mathbf{c}_i$, thereby reducing the memory footprint of intermediate representations. This design significantly reduces the computational cost while preserving essential structural information. More importantly, the latent-space formulation enables efficient modeling of long-range cross-box interactions by allowing each box to attend to globally aggregated latent representations rather than individual box features. This provides a compact yet expressive mechanism to capture global structural dependencies in cryo-EM density maps, which are often spatially sparse and noisy. Based on the latent representations, the attention output at position $i$ is computed as
\begin{equation}
\mathbf{o}_i = \sum_{j=1}^{L} \text{softmax}_j \left(
\frac{(\mathbf{W}_q \tilde{\mathbf{E}}_i^{\text{box}})^\top (\mathbf{W}_k \mathbf{c}_j)}{\sqrt{d}}
\right)(\mathbf{W}_v \mathbf{c}_j),
\end{equation}
where $\mathbf{W}_q$, $\mathbf{W}_k$, and $\mathbf{W}_v$ are learnable projection matrices. Final map representation is obtained as
\begin{equation}
\tilde{\mathbf{E}}^{\text{map}} = f_{\text{MLA}}(\mathbf{E}^{\text{map}}),
\end{equation}
where $f_{\text{MLA}}(\cdot)$ denotes the proposed architecture, and $\tilde{\mathbf{E}}^{\text{map}} \in \mathbb{R}^{L \times 240}$. The detailed design of the MLA-based Transformer architecture is provided in Appendix~\ref{MLA-based Transformer architecture}. This design enables efficient and scalable modeling of cross-box interactions in cryo-EM density maps.

\subsection{Multi-task Pretraining}
As illustrated in Fig.~\ref{fig:framework} (c), we design three pretraining tasks to enable the model to learn generalizable representations from cryo-EM density maps and protein sequences, including cross-modal prediction, masked map reconstruction, and modality classification. These tasks are not independent but are designed to complement each other from different perspectives. Specifically, cross-modal prediction focuses on aligning global semantic representations between sequence and map modalities, masked map reconstruction emphasizes learning fine-grained local structural patterns, while modality classification enforces discriminative feature learning across modalities. Together, they provide a unified training objective that jointly captures global alignment, local structure, and modality-aware representations, leading to more robust and generalizable features. Detailed formulations of these pretraining tasks are provided in Appendix~\ref{appendix:pretraining task}.

\textbf{Cross-modal Prediction.}
This task aims to align sequence and map representations while enabling the model to predict map features from sequence information alone. By learning this cross-modal mapping, the model can infer map representations from sequences, enabling downstream prediction without requiring cryo-EM density maps. This facilitates effective knowledge transfer between modalities and improves robustness when one modality is missing.

\textbf{Masked Map Reconstruction.}
This task enhances the model’s ability to capture local structural patterns by randomly masking a subset of input boxes and reconstructing the corresponding density representations within the learned feature space. By enforcing reconstruction from partial observations, the model is encouraged to learn fine-grained structural dependencies and contextual relationships across neighboring regions.

\begin{table*}[!t]
\centering
\caption{
Performance comparison on residue-level prediction tasks.
Seq and Str denote sequence and structure, respectively.
CryoProt uses maps only during pretraining. The best results are in bold, and the second-best results are underlined.
}
\label{tab:comparison_results1}

\setlength{\tabcolsep}{3.5pt}
\renewcommand{\arraystretch}{1.0}

\begin{tabular}{c|ccc|ccc | c|ccc|ccc}
\toprule

\multicolumn{7}{c|}{\textbf{\textit{A. Protein flexibility prediction (residue-level)}}}
&
\multicolumn{7}{c}{\textbf{\textit{B. Active site identification (residue-level)}}} \\

\textbf{Model}
& \textbf{Seq} & \textbf{Str} & \textbf{Map}
& \textbf{RMSE $\downarrow$} & \textbf{PC $\uparrow$} & \textbf{SC $\uparrow$}
& \textbf{Model}
& \textbf{Seq} & \textbf{Str} & \textbf{Map}
& \textbf{AUPRC $\uparrow$} & \textbf{MCC $\uparrow$} & \textbf{FPR $\downarrow$} \\
\midrule

\rowcolor{gray!15}
\multicolumn{7}{c}{\textit{Domain-specific models}}
&
\multicolumn{7}{c}{\textit{Domain-specific models}} \\

DEFMap
& \xmark & \xmark & \cmark
& 0.550 & 0.203 & 0.033
& Deep-ProBind
& \cmark & \cmark & \xmark
& 0.539 & 0.537 & 0.032 \\

RMSF-net
& \xmark & \cmark & \cmark
& \underline{0.495} & 0.273 & 0.222
& MMSite
& \cmark & \xmark & \xmark
& 0.643 & 0.592 & 0.028 \\

ResaPred
& \cmark & \xmark & \xmark
& 0.527 & 0.186 & 0.014
& M$^3$Site
& \cmark & \cmark & \xmark
& 0.735 & 0.644 & 0.024 \\

Flexpert-3D
& \cmark & \cmark & \xmark
& 0.501 & 0.201 & 0.208
& DeepProSite
& \cmark & \cmark & \xmark
& 0.593 & 0.561 & 0.030 \\

\rowcolor{gray!15}
\multicolumn{7}{c}{\textit{Pretrained models}}
&
\multicolumn{7}{c}{\textit{Pretrained models}} \\

ESM2-150M
& \cmark & \xmark & \xmark
& 0.499 & 0.296 & 0.279
& ESM2-150M
& \cmark & \xmark & \xmark
& 0.705 & 0.612 & 0.028 \\

OmegaFold
& \cmark & \cmark & \xmark
& 0.497 & 0.245 & 0.252
& OmegaFold
& \cmark & \cmark & \xmark
& 0.719 & 0.628 & 0.026 \\

ESM3
& \cmark & \xmark & \xmark
& 0.619 & 0.248 & 0.260
& ESM3
& \cmark & \xmark & \xmark
& 0.653 & 0.612 & 0.027 \\

SaProt
& \cmark & \cmark & \xmark
& 0.498 & 0.304 & 0.307
& SaProt
& \cmark & \cmark & \xmark
& 0.645 & 0.601 & 0.025 \\

ProstT5
& \cmark & \cmark & \xmark
& 0.515 & \textbf{0.327} & \underline{0.314}
& ProstT5
& \cmark & \cmark & \xmark
& \underline{0.744} & 0.658 & \textbf{0.016} \\

ProST
& \cmark & \xmark & \xmark
& 0.503 & 0.307 & 0.307
& ProST
& \cmark & \xmark & \xmark
& 0.742 & \underline{0.659} & 0.022 \\

\rowcolor[HTML]{F2F2FF}
\textbf{CryoProt}
& \cmark & \xmark & \cmark
& \textbf{0.484} & \underline{0.310} & \textbf{0.315}
& \textbf{CryoProt}
& \cmark & \xmark & \cmark
& \textbf{0.746} & \textbf{0.669} & \underline{0.021} \\

\bottomrule
\end{tabular}
\end{table*}

\textbf{Modality Classification.}
Inspired by prior studies \citep{xiong2026multi, ganin2015unsupervised, wang2023multi}, this task predicts the modality type (e.g., sequence or map) of each representation to improve modality-specific feature learning and enhance distinguishability between modalities. It further regularizes the shared representation space and prevents feature collapse across modalities.

We assign a balancing coefficient $\lambda_i$ to each pretraining task to control the contribution of different loss terms, and adopt an uncertainty-based weighting scheme \citep{kendall2018multi, kendall2017uncertainties} to automatically determine these coefficients. This dynamic weighting strategy adaptively balances the learning difficulty of different tasks, ensuring stable optimization and preventing domination by any single objective. The overall training objective is defined as
\begin{equation}
\mathcal{L} = \sum_{i \in \{\text{cross}, \text{mask}, \text{class}\}} \left( \frac{1}{2\sigma_i^2} \mathcal{L}_i + \log \sigma_i \right),
\end{equation}
where $\mathcal{L}$ denotes the total pretraining loss, and $\mathcal{L}_i$ represents the task-specific loss for the $i$-th pretraining objective. $\sigma_i$ is a learnable parameter representing the uncertainty of the $i$-th task, and the corresponding coefficients satisfy $\lambda_i = \frac{1}{2\sigma_i^2}$.

\section{Experiments and Results}
\label{result}

\subsection{Experimental Setup}
\label{experimental_setup}

All experiments are conducted on a Linux server with 25 CPU cores and two NVIDIA GeForce RTX 5090 GPUs with 64 GB of VRAM. We first pretrain the CryoProt model on our pretraining cryo-EM dataset and then fine-tune it on four downstream tasks, with performance evaluated via five-fold cross-validation to ensure robustness and reliability. During pretraining, the model learns generalizable representations from unlabeled Cryo-EM data, which are subsequently adapted to task-specific objectives in the fine-tuning stage. Detailed training configurations, hyperparameter settings, fine-tuning procedures, baseline methods, and evaluation metrics are provided in Appendix~\ref{appendix:experiment details}.

\subsection{Comparison Results}

\begin{table*}[!t]
\centering
\caption{
Performance comparison on protein-level prediction tasks.
Seq and Str denote sequence and structure, respectively.
CryoProt uses maps only during pretraining. The best results are in bold, and the second-best results are underlined.
}
\label{tab:comparison_results2}

\setlength{\tabcolsep}{4pt}
\renewcommand{\arraystretch}{1.0}

\begin{tabular}{c|ccc|ccc | c|ccc|ccc}
\toprule

\multicolumn{7}{c|}{\textbf{\textit{C. Binding affinity prediction (protein-level)}}}
&
\multicolumn{7}{c}{\textbf{\textit{D. $\Delta\Delta G$ prediction (protein-level)}}} \\

\textbf{Model}
& \textbf{Seq} & \textbf{Str} & \textbf{Map}
& \textbf{RMSE $\downarrow$} & \textbf{PC $\uparrow$} & \textbf{SC $\uparrow$}
& \textbf{Model}
& \textbf{Seq} & \textbf{Str} & \textbf{Map}
& \textbf{RMSE $\downarrow$} & \textbf{PC $\uparrow$} & \textbf{SC $\uparrow$} \\
\midrule

\rowcolor{gray!15}
\multicolumn{7}{c}{\textit{Domain-specific models}}
&
\multicolumn{7}{c}{\textit{Domain-specific models}} \\

Bind-ddG
& \cmark & \cmark & \xmark
& 0.653 & 0.392 & 0.384
& PPIformer
& \xmark & \cmark & \xmark
& 1.327 & 0.598 & \underline{0.468} \\

GearBind
& \xmark & \cmark & \xmark
& 0.502 & \underline{0.495} & \underline{0.508}
& ProBASS
& \cmark & \cmark & \xmark
& 1.183 & 0.591 & 0.433 \\

PPIgraphomer
& \cmark & \cmark & \xmark
& 0.731 & 0.252 & 0.336
& DGCddG
& \xmark & \cmark & \xmark
& 1.463 & 0.352 & 0.309 \\

Island
& \cmark & \xmark & \xmark
& 0.915 & 0.108 & 0.076
& MFFN
& \cmark & \cmark & \xmark
& 1.518 & 0.258 & 0.334 \\

\rowcolor{gray!15}
\multicolumn{7}{c}{\textit{Pretrained models}}
&
\multicolumn{7}{c}{\textit{Pretrained models}} \\

ESM2-150M
& \cmark & \xmark & \xmark
& 0.546 & 0.386 & 0.402
& ESM2-150M
& \cmark & \xmark & \xmark
& 1.219 & 0.548 & 0.369 \\

OmegaFold
& \cmark & \cmark & \xmark
& 0.521 & 0.462 & 0.471
& OmegaFold
& \cmark & \cmark & \xmark
& 1.178 & 0.571 & 0.402 \\

ESM3
& \cmark & \xmark & \xmark
& 0.629 & 0.230 & 0.221
& ESM3
& \cmark & \xmark & \xmark
& 1.161 & 0.491 & 0.415 \\

SaProt
& \cmark & \cmark & \xmark
& 0.493 & 0.456 & 0.480
& SaProt
& \cmark & \cmark & \xmark
& \underline{1.121} & 0.513 & 0.429 \\

ProstT5
& \cmark & \cmark & \xmark
& 0.512 & 0.413 & 0.446
& ProstT5
& \cmark & \cmark & \xmark
& 1.126 & \textbf{0.622} & 0.439 \\

ProST
& \cmark & \xmark & \xmark
& \underline{0.492} & 0.450 & 0.459
& ProST
& \cmark & \xmark & \xmark
& 1.126 & 0.506 & 0.432 \\

\rowcolor[HTML]{F2F2FF}
\textbf{CryoProt}
& \cmark & \xmark & \cmark
& \textbf{0.486} & \textbf{0.524} & \textbf{0.527}
& \textbf{CryoProt}
& \cmark & \xmark & \cmark
& \textbf{1.036} & \underline{0.611} & \textbf{0.488} \\

\bottomrule
\end{tabular}
\end{table*}

The comparison results are shown in Table~\ref{tab:comparison_results1} and Table~\ref{tab:comparison_results2}. Overall, CryoProt achieves higher mean performance on most metrics across four downstream tasks. Although some improvements fall within the standard deviations of competing methods, its competitive performance across diverse tasks supports the effectiveness and transferability of the learned representations. Specifically, for protein flexibility prediction, CryoProt
achieves the lowest RMSE and competitive correlation scores among sequence- and structure-based pretrained models. This indicates that cryo-EM density maps provide complementary structural dynamics information for modeling protein flexibility.
Compared with density map-based methods such as RMSF-net, CryoProt achieves better performance, potentially benefiting from explicit cross-box interactions that capture global structural dependencies.
Although CryoProt adopts ESM2-150M as its Sequence Encoder, it outperforms ESM2-150M alone across all three flexibility metrics, suggesting that the gains also arise from learning structural information in cryo-EM maps. More importantly, the contribution of this work is not limited to numerical gains. We propose a protein pretraining framework that exploits experimental cryo-EM density maps for downstream property prediction, in which cross-box interactions are explicitly modeled to learn transferable protein representations.

\subsection{Ablation Study}
\label{ablation study}

\begin{figure}[!t]
\centering
\includegraphics[width=\linewidth]{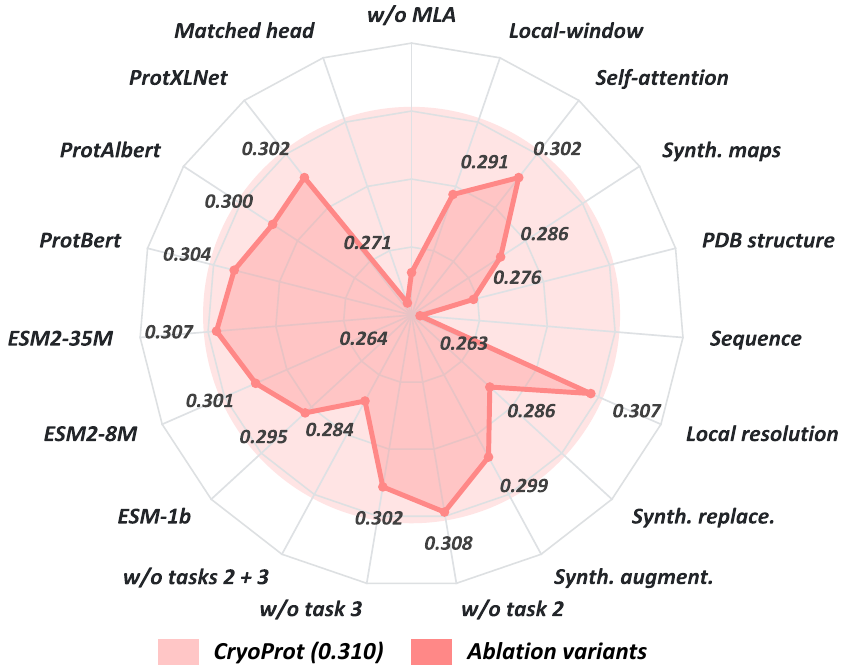}
\caption{PC of ablation variants on protein flexibility prediction. The pale pink circle indicates CryoProt (PC = 0.310), while the coral profile shows the ablation results.}
\label{fig:ablation}
\end{figure}

We conduct ablation studies on cross-box interactions, cryo-EM maps, local resolution, pretraining tasks, and sequence encoder variants.
Figure~\ref{fig:ablation} summarizes the ablation results for protein flexibility prediction.
Additional analyses of the ablation variants and results on the other downstream tasks are provided in Appendix~\ref{appendix:ablation study}.

\textbf{Box Interactions.}
Removing MLA disables cross-box interactions and reduces PC from 0.310 to 0.271. Replacing MLA with local-window attention or self-attention yields PC values of 0.291 and 0.302, respectively. For local-window attention, each box attends to its 30 spatially nearest neighboring boxes based on Euclidean distances between box centers. Both alternatives outperform the variant without cross-box interactions, but MLA achieves the best performance, suggesting that its latent global interactions help capture dependencies among boxes.

\textbf{Pretraining Tasks.}
We remove masked map reconstruction, modality classification, or both to assess their contributions. Cross-modal prediction is retained in all variants because it enables map representation inference from sequences when density maps are unavailable. Removing the auxiliary tasks individually reduces PC from 0.310 to 0.308 and 0.302, while removing both further reduces PC to 0.284. The larger decline under joint removal suggests complementary benefits from the two auxiliary objectives.

\subsection{Parameter Sensitivity Analysis}

\begin{figure}[!h]
\centering
\includegraphics[width=\linewidth]{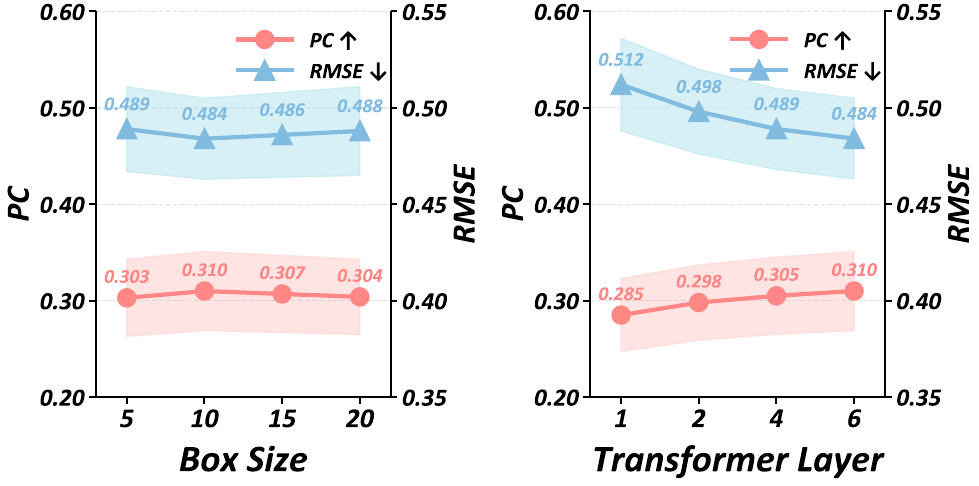}
\caption{Parameter sensitivity analysis for box size and Transformer layers.}
\label{fig:parameter_sensitivity_analysis}
\end{figure}

We analyze the sensitivity of box size and Transformer depth on protein flexibility prediction, as shown in Fig.~\ref{fig:parameter_sensitivity_analysis}. Additional sensitivity analyses for the attention heads and mask ratio are provided in Appendix~\ref{appendix:parameter_sensitivity}.

\begin{figure*}[!hb] 
\centering 
\includegraphics[width=\linewidth]{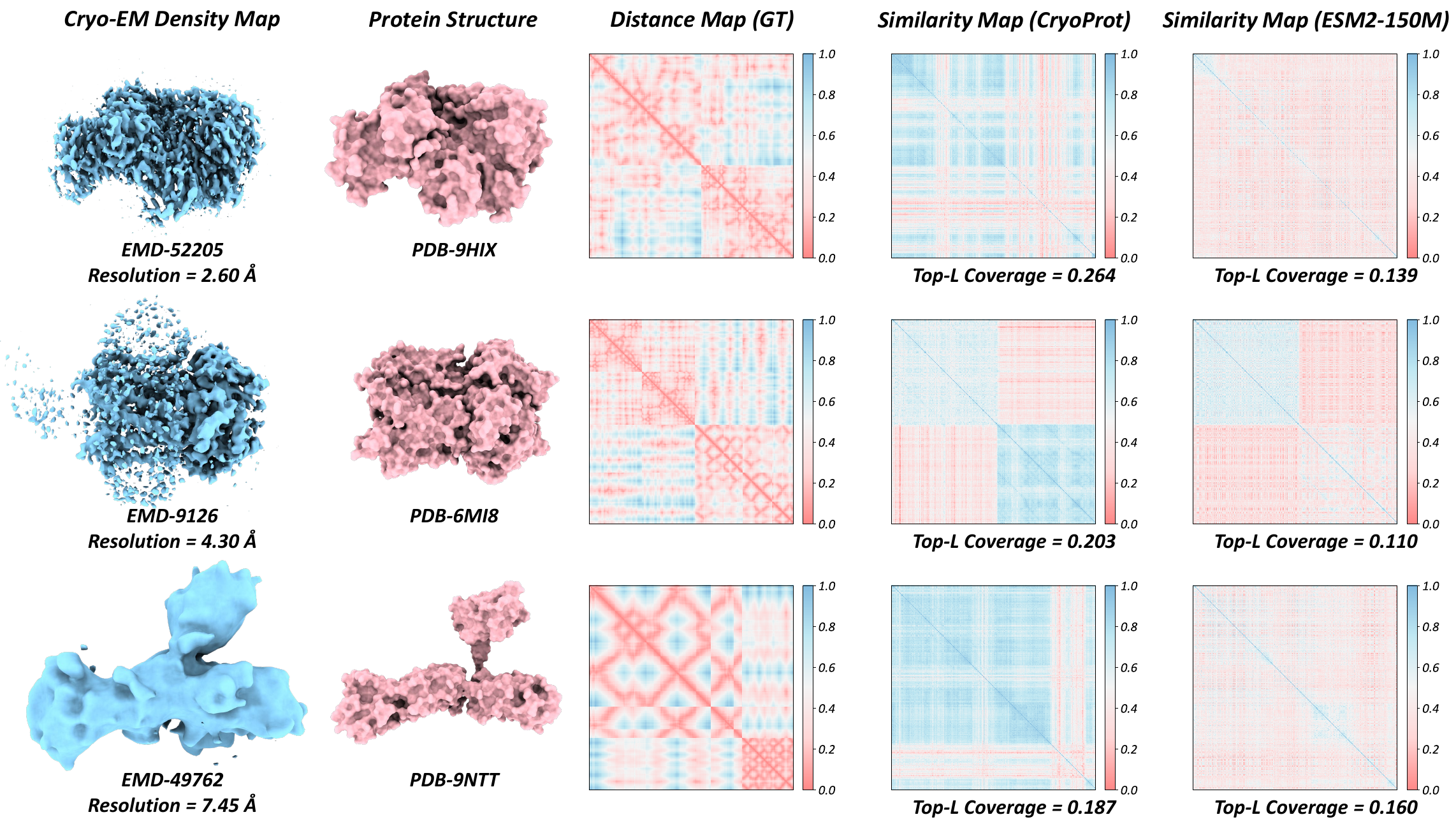} 
\caption{Visualization and comparison of residue distance maps and embedding similarity maps generated by CryoProt and ESM2 across different resolution ranges.} 
\label{fig:case_on_top} 
\end{figure*}

\textbf{Box Size.} Performance remains relatively stable across different settings, indicating that CryoProt is not highly sensitive to spatial granularity. The best overall performance is achieved at a box size of 10, while larger boxes substantially increase memory and computational costs, with a box size of 20 approaching the hardware limit. 

\textbf{Transformer Layer.} Performance generally improves as the number of layers increases, suggesting that deeper architectures can better capture cross-box dependencies. Due to computational constraints, we therefore set the maximum depth to 6 layers.

\subsection{Case Study}
\label{case}

\begin{figure*}[!t]
\centering
\includegraphics[width=\linewidth]{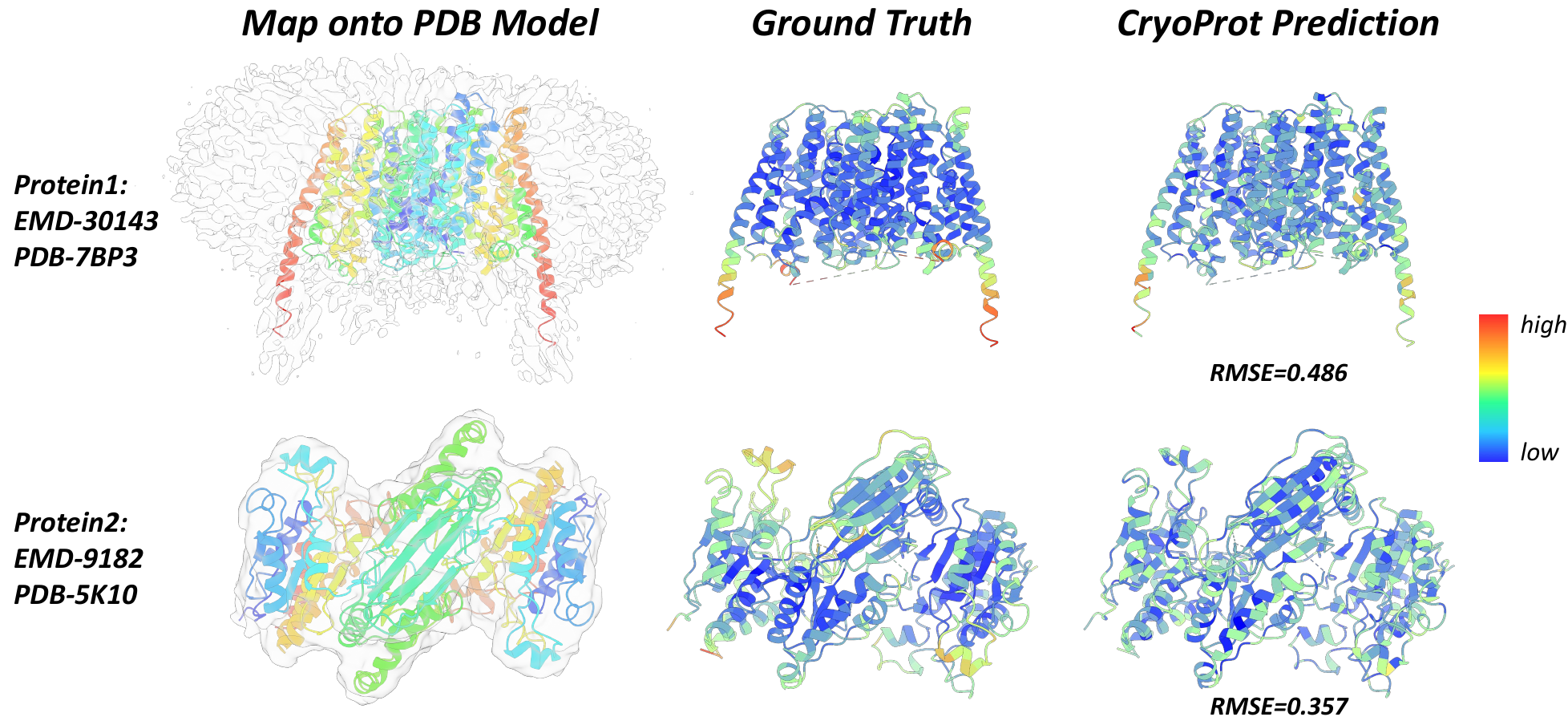}
\caption{Case study on protein flexibility prediction. The ground-truth and predicted flexibility are visualized on protein structures.}
\label{fig:case_on_rmsf}
\end{figure*}

\textbf{Case on Top-$L$ coverage.}
Due to the inherent resolution heterogeneity in cryo-EM density maps
\citep{cushing2024high, vilas2020measuring, ramirez2020automatic},
we evaluate whether CryoProt can capture structural relationships under varying resolution conditions. Specifically, we randomly select 40 proteins from each of three resolution ranges (2–4 Å, 4–6 Å, and 6–8 Å), yielding 120 proteins in total, and compute their ground-truth residue distance maps from the corresponding native structures. We then construct residue
similarity maps based on embeddings from CryoProt and ESM2 and assess
their alignment with the distance maps using the Top-$L$ coverage metric
\citep{li2019respre, xiong2017deep}.
As summarized in Table~\ref{tab:topl_results}, CryoProt consistently outperforms different ESM2 variants across all resolution ranges, indicating that it can implicitly capture spatial relationships without explicit structural
supervision and remains robust under low-resolution conditions. As shown in Fig.~\ref{fig:case_on_top}, representative proteins from different resolution ranges consistently show that the similarity maps produced by CryoProt better align with the ground-truth residue distance maps than those produced by ESM2, further demonstrating the ability of CryoProt to capture residue-level spatial relationships from cryo-EM density information. Detailed definitions and interpretation of the Top-$L$ coverage metric, together with per-protein results, are provided in Appendix~\ref{appendix:case_on_topL}.

\begin{table}[!h]
\centering
\caption{
Top-$L$ coverage comparison. The best results are in bold, and the second-best results are underlined. $\Delta$ denotes the relative improvement of CryoProt over the second-best model.
}
\label{tab:topl_results}
\renewcommand{\arraystretch}{1}
\small
\setlength{\tabcolsep}{3pt}
\begin{tabular}{c c c c c c}
\toprule
\textbf{Resolution Range}
& \textbf{8M}
& \textbf{35M}
& \textbf{150M}
& \textbf{CryoProt}
& \textbf{$\Delta$} \\
\midrule
2--4 Å
& 0.135
& 0.141
& \underline{0.147}
& \textbf{0.166}
& $\uparrow$ 12.93\% \\

4--6 Å
& 0.120
& 0.127
& \underline{0.132}
& \textbf{0.159}
& $\uparrow$ 20.45\% \\

6--8 Å
& 0.101
& 0.108
& \underline{0.112}
& \textbf{0.128}
& $\uparrow$ 14.29\% \\

Mean
& 0.119
& 0.125
& \underline{0.130}
& \textbf{0.151}
& $\uparrow$ 16.15\% \\
\bottomrule
\end{tabular}
\end{table}

\textbf{Case on protein flexibility prediction.}
We further present qualitative results on two representative proteins, identified by their PDB IDs 7BP3 (EMD-30143, \textit{human monocarboxylate transporter}) and 5K10 (EMD-9182, \textit{isocitrate dehydrogenase}), which are associated with disease-related biological processes. The ground-truth and predicted flexibility are visualized on the 3D structures using color mapping with ChimeraX software\footnote{\url{https://www.cgl.ucsf.edu/chimerax}}, as shown in Fig.~\ref{fig:case_on_rmsf}. CryoProt achieves RMSE values of 0.486 and 0.357 on two proteins, respectively, and the predicted patterns exhibit good consistency with the ground-truth distributions. These results demonstrate that our model can effectively capture residue-level flexibility and generalize well to downstream tasks. Additional visualization results are provided in Appendix~\ref{appendix:more_visualization}.

\section{Discussion}
\label{discussion}

In this work, we propose CryoProt, a pretraining framework that leverages cryo-EM density maps to learn protein representations. By introducing an MLA-based Transformer architecture to model interactions across local boxes, together with a multi-task pretraining strategy, CryoProt consistently outperforms state-of-the-art methods across multiple downstream tasks. Through case studies on Top-$L$ coverage and visualization, we further demonstrate that CryoProt maintains strong performance across proteins with varying resolution levels. Notably, the model can implicitly capture protein spatial relationships using only sequence information and cryo-EM density maps, without requiring explicit structural supervision. However, CryoProt is currently limited to protein density maps for pretraining. Future work will explore extending this framework to other biomolecular systems, such as RNA and viral proteins. Moreover, incorporating generative modeling to jointly learn protein structures and cryo-EM density maps may provide a promising direction to better address resolution heterogeneity and noise in density data.

\section*{AI Use Statement}

We used OpenAI Codex to assist with grammar checking and LaTeX formatting. It was not used to generate research data, design or run experiments, implement the reported method, formulate mathematical claims, assist with proofs, or compute the reported results. All AI-assisted edits were checked against the source manuscript and reported tables. The authors take full responsibility for the final text, claims, and artifacts.

\section*{Impact Statement}

CryoProt aims to advance protein representation learning by incorporating information from experimental cryo-EM density maps. This work may support studies of protein function, with potential benefits for biomedical research and drug discovery. Predictions should be experimentally validated before informing biological or therapeutic decisions. CryoProt should be used with caution, as protein modeling tools could potentially be misused for harmful biological applications, including the design or modification of viruses. Users should follow applicable biosafety and biosecurity requirements.

\nocite{langley00}

\bibliography{ref}
\bibliographystyle{icml2027}


\clearpage
\appendix
\section*{Appendix}

\section{MLA-based Transformer Architecture}
\label{MLA-based Transformer architecture}

The cryo-EM density map is represented as a sequence of box embeddings
\begin{equation}
\mathbf{E}^{\text{map}} = \{\tilde{\mathbf{E}}_1^{\text{box}}, \tilde{\mathbf{E}}_2^{\text{box}}, \dots, \tilde{\mathbf{E}}_L^{\text{box}}\},
\end{equation}
where $\tilde{\mathbf{E}}_i^{\text{box}} \in \mathbb{R}^{d}$ denotes the embedding of the $i$-th box, and $L$ denotes the sequence length, which is equal to the number of boxes. The architecture consists of $L_{\text{layer}}$ stacked Transformer layers, each composed of a Multi-Head Latent Attention (MLA) module followed by a feed-forward network (FFN). At the $l$-th layer, the input representations are denoted as $\mathbf{H}^{(l)} = \{\mathbf{h}_1^{(l)}, \dots, \mathbf{h}_L^{(l)}\}$. For each token, MLA first projects the input into a lower-dimensional latent space
\begin{equation}
\mathbf{c}_i^{(l)} = \mathbf{W}_{D}^{(l)} \mathbf{h}_i^{(l)},
\end{equation}
where $\mathbf{W}_{D}^{(l)} \in \mathbb{R}^{d_c \times d}$ is a learnable projection matrix and $d_c \ll d$. This latent projection compresses the input representations, enabling more efficient attention computation while retaining essential structural information. The latent representations are used to construct keys and values, while queries are obtained from the original input
\begin{equation}
\mathbf{q}_i^{(l)} = \mathbf{W}_{Q}^{(l)} \mathbf{h}_i^{(l)}, \quad
\mathbf{k}_i^{(l)} = \mathbf{W}_{K}^{(l)} \mathbf{c}_i^{(l)}, \quad
\mathbf{v}_i^{(l)} = \mathbf{W}_{V}^{(l)} \mathbf{c}_i^{(l)},
\end{equation}
where $\mathbf{W}_{Q}^{(l)}$, $\mathbf{W}_{K}^{(l)}$, and $\mathbf{W}_{V}^{(l)}$ are learnable parameters. In practice, multiple attention heads are used to capture diverse interaction patterns, and the outputs from different heads are concatenated and linearly projected. The attention output at position $i$ is computed by aggregating information from all tokens in the sequence
\begin{equation}
\mathbf{o}_i^{(l)} = \sum_{j=1}^{L} \text{softmax}_j \left(
\frac{(\mathbf{q}_i^{(l)})^\top \mathbf{k}_j^{(l)}}{\sqrt{d_c}}
\right)\mathbf{v}_j^{(l)},
\end{equation}
which enables each box to attend to global contextual information through the latent representations. Compared with standard attention, this formulation avoids directly computing pairwise interactions in the original feature space, significantly reducing computational overhead. A residual connection is applied to combine the input and attention output
\begin{equation}
\hat{\mathbf{H}}^{(l)} = \mathbf{H}^{(l)} + \mathbf{o}^{(l)},
\end{equation}
where $\mathbf{o}^{(l)} = \{\mathbf{o}_1^{(l)}, \dots, \mathbf{o}_L^{(l)}\}$. The representations are then fed into a feed-forward network (FFN), followed by another residual connection
\begin{equation}
\mathbf{H}^{(l+1)} = \hat{\mathbf{H}}^{(l)} + \text{FFN}(\hat{\mathbf{H}}^{(l)}),
\end{equation}
where the FFN is implemented as a two-layer MLP with non-linear activation. After $L_{\text{layer}}$ layers, a linear projection is applied to align the feature dimension with the Sequence Encoder
\begin{equation}
\tilde{\mathbf{E}}^{\text{map}} = \mathbf{H}^{(L_{\text{layer}})} \mathbf{W}_o,
\end{equation}
where $\mathbf{W}_o$ is a learnable projection matrix, resulting in $\tilde{\mathbf{E}}^{\text{map}} \in \mathbb{R}^{L \times 240}$.

Compared with standard multi-head attention, MLA projects box embeddings into a lower-dimensional latent space, reducing the memory overhead of intermediate feature representations while retaining quadratic attention complexity. This design preserves the ability to model long-range dependencies while improving memory efficiency. Such a property is particularly suitable for cryo-EM density maps, where modeling interactions between spatially distant boxes is essential but challenging due to the large number of boxes and the sparsity of meaningful signals.

\begin{figure*}[!t]
\centering
\begin{lstlisting}[style=paperbash]
for dataset in RMSFNet ProTAD HER2 SKEMPIv2; do
    for identity in 0.30 0.50 0.70 0.90; do
        mmseqs easy-search \
            ${dataset}.fasta \
            pretrain.fasta \
            ${dataset}_${identity}.m8 \
            tmp \
            --min-seq-id ${identity} \
            -c 0.80 \
            --cov-mode 0 \
            --format-output query,target,fident,qcov,tcov
    done
done
\end{lstlisting}
\caption{MMseqs2 commands for sequence similarity analysis
between downstream datasets and the pretraining dataset.}
\label{fig:mmseqs}
\end{figure*}

\section{Multi-task Pretraining Objectives}
\label{appendix:pretraining task}

\textbf{Cross-modal Prediction.}
Given the sequence embedding $\tilde{\mathbf{E}}^{\text{seq}} \in \mathbb{R}^{L \times d}$ and map embedding $\tilde{\mathbf{E}}^{\text{map}} \in \mathbb{R}^{L \times d}$, we employ a learnable projection function to predict map representations from sequence features. Specifically, each sequence token embedding is mapped into the map representation space as
\begin{equation}
\hat{\mathbf{E}}^{\text{map}}_i = f_{\text{proj}}(\tilde{\mathbf{E}}^{\text{seq}}_i),
\end{equation}
where $f_{\text{proj}}(\cdot)$ denotes a learnable mapping function implemented as a multi-layer perceptron (MLP). This task encourages alignment between sequence and map modalities by minimizing the discrepancy between predicted and ground-truth map embeddings. The cross-modal prediction loss is defined as
\begin{equation}
\mathcal{L}_{\text{cross}} = \frac{1}{L} \sum_{i=1}^{L} 
\left\| \hat{\mathbf{E}}^{\text{map}}_i - \tilde{\mathbf{E}}^{\text{map}}_i \right\|_2^2.
\end{equation}
This objective enables the model to infer structural representations from sequence information, facilitating downstream prediction when cryo-EM data is unavailable.

\textbf{Masked Map Reconstruction.}
To enhance the model’s ability to capture local structural dependencies, we randomly mask a subset of box representations during training. Let $\mathcal{M}$ denote the set of masked positions sampled from the input sequence, where each position is masked with a fixed probability. The model is then trained to reconstruct the original box embeddings based on the remaining visible context. The reconstruction loss is defined as
\begin{equation}
\mathcal{L}_{\text{mask}} = \frac{1}{|\mathcal{M}|} \sum_{i \in \mathcal{M}} 
\left\| \hat{\mathbf{E}}^{\text{box}}_i - \tilde{\mathbf{E}}^{\text{box}}_i \right\|_2^2,
\end{equation}
where $\hat{\mathbf{E}}^{\text{box}}_i$ denotes the reconstructed representation obtained from the model. This objective encourages the model to learn fine-grained local structures and contextual relationships across neighboring regions.

\textbf{Modality Classification.}
To enhance modality-specific feature learning, we introduce a token-level binary classifier that distinguishes between the two modalities. A lightweight classification head outputs the probability of each token belonging to one of the two modalities using a sigmoid function. The classification loss is
\begin{equation}
\mathcal{L}_{\text{class}}
= -\frac{1}{L}\sum_{i=1}^{L}
\left[
y_i \log p_i
+
(1-y_i)\log(1-p_i)
\right],
\end{equation}
where $y_i \in {0,1}$ denotes the modality label of token $i$, and $p_i$ is the predicted probability that token $i$ belongs to the positive modality. This objective encourages the learned representations to retain modality-specific information.

\section{Dataset Details}
\label{appendix:dataset}

\textbf{Pretraining dataset.} The pretraining data are collected from EMDB \citep{lawson2016emdatabank}, which provides a large number of experimentally determined cryo-EM density maps. Following the data selection protocol of DEFMap \citep{matsumoto2021extraction}, we retain protein density maps with resolutions ranging from 2~\AA\ to 4~\AA, where relatively high-resolution maps provide more reliable structural details for representation learning. In addition, we restrict the protein sequence length to be less than 1024, following the setting of ESM2, to ensure compatibility with the Sequence Encoder. To reduce the variability across different experimental conditions (e.g., microscope settings, noise levels, and local resolution), we adopt a normalization strategy inspired by EModelX \citep{chen2024protein}. Given a raw density map $M \in \mathbb{R}^{w \times h \times d}$, we first align its coordinate system with the corresponding PDB structure and resample it to a unified voxel size of $1 \times 1 \times 1$~\AA\ via trilinear interpolation, resulting in a processed map $M' \in \mathbb{R}^{w' \times h' \times d'}$. The normalized map $N$ is then computed as
\begin{equation}
N_{xyz} =
\left\{
\begin{aligned}
&0, && M'_{xyz} < M'_{\text{med}} \\
&\dfrac{M'_{xyz} - M'_{\text{med}}}{M'_{\text{top1}} - M'_{\text{med}}}, && M'_{\text{med}} \leq M'_{xyz} < M'_{\text{top1}} \\
&1, && M'_{xyz} \geq M'_{\text{top1}}
\end{aligned}
\right.
\end{equation}

where $(x, y, z)$ denotes the voxel coordinate, $M'_{\text{med}}$ is the median density value, and $M'_{\text{top1}}$ denotes the top 1\% density value. This normalization constrains all voxel intensities to the range $[0,1]$, reducing noise effects and improving training stability.

\textbf{Downstream datasets.} We evaluate CryoProt on four representative downstream tasks. The protein flexibility prediction dataset is obtained from RMSFNet \citep{song2024accurate}, which provides residue-level flexibility annotations. The active site identification dataset is derived from ProTAD \citep{ouyang2024mmsite}, containing annotated catalytic residues. The protein--protein binding affinity prediction task uses the HER2 dataset \citep{shanehsazzadeh2023unlocking}, while $\Delta\Delta G$ prediction is evaluated on the SKEMPI v2 dataset \citep{jankauskaite2019skempi}. Following the pretraining setting, we further filter all downstream datasets by restricting protein sequence lengths to less than 1024, and only retain samples that satisfy this criterion. In addition, we remove any proteins that overlap with the pretraining dataset to prevent potential data leakage and ensure a fair evaluation. These tasks span both residue-level and protein-level prediction settings, where the first two tasks are defined at the residue level and the latter two are at the protein level, enabling a comprehensive evaluation of the learned representations. 

\textbf{Similarity analysis.}
Although proteins with exact overlap between the pretraining and downstream datasets are removed, proteins with similar sequences may still exist across the datasets. To further assess potential sequence-level data leakage, we perform an additional similarity analysis using MMseqs2. Specifically, each of the four downstream datasets is independently searched against the same pretraining set at sequence identity thresholds of 30\%, 50\%, 70\%, and 90\%, with at least 80\% alignment coverage for both query and target sequences.For reproducibility, an example of the MMseqs2 commands used in our analysis is shown in Figure~\ref{fig:mmseqs}. The results are summarized in Table~\ref{tab:sequence_similarity}. Overall, only a small fraction of downstream proteins exhibit high sequence similarity to proteins in the pretraining set. At the 90\% identity threshold, the proportions are only 0.30\%, 0.33\%, 0.00\%, and 2.77\% for RMSFNet, ProTAD, HER2, and SKEMPI v2, respectively. These results suggest that the observed downstream performance is unlikely to be primarily driven by sequence-level overlap with the pretraining data. We focus on sequence similarity rather than exhaustive structure-level comparison. Structural similarity is harder to define consistently because proteins may differ in conformation, chain composition, and structural coverage, and simple metrics such as RMSD or TM-score may not directly indicate data leakage. We therefore adopt sequence similarity as a standardized and scalable criterion for assessing potential data leakage.

\begin{table}[!h]
\centering
\caption{Sequence similarity analysis between downstream and pretraining datasets.}
\label{tab:sequence_similarity}
\small
\setlength{\tabcolsep}{3pt}
\renewcommand{\arraystretch}{1}
\begin{tabular}{ccccc}
\toprule
\textbf{Dataset} &
$\boldsymbol{\geq}$\textbf{30\%} &
$\boldsymbol{\geq}$\textbf{50\%} &
$\boldsymbol{\geq}$\textbf{70\%} &
$\boldsymbol{\geq}$\textbf{90\%} \\
\midrule
RMSFNet   & 4 (1.2\%)   & 3 (0.9\%)   & 1 (0.3\%)   & 1 (0.3\%) \\
ProTAD    & 63 (1.0\%)  & 43 (0.7\%)  & 33 (0.5\%)  & 21 (0.3\%) \\
HER2      & 5 (1.2\%)   & 2 (0.5\%)   & 0 (0.0\%)   & 0 (0.0\%) \\
SKEMPI v2 & 275 (3.8\%) & 274 (3.8\%) & 272 (3.8\%) & 200 (2.8\%) \\
\bottomrule
\end{tabular}
\end{table}

\textbf{Data splitting.} To avoid potential protein data leakage \citep{alquraishi2019proteinnet, vskrhak2025cryptobench}, all datasets are split at the protein level rather than at the residue level (i.e., residues from the same protein are kept within the same fold rather than being split across folds), and a five-fold cross-validation strategy is employed for robust evaluation.

\section{More Experiment Details}
\label{appendix:experiment details}

\subsection{Hyperparameter settings}
\label{appendix:hyperparameter settings}

The detailed hyperparameter settings used in both pretraining and fine-tuning are summarized in Table~\ref{tab:hyperparameters}. The listed values of $\lambda_1$, $\lambda_2$, and $\lambda_3$ are their initial values, and these weights are subsequently learned through an uncertainty-based weighting scheme to balance the training objectives.

\begin{table}[h]
\centering
\caption{Training hyperparameters of CryoProt.}
\setlength{\tabcolsep}{1pt}
\renewcommand{\arraystretch}{1}
\label{tab:hyperparameters}
\begin{tabular}{c c | c c}
\toprule
\textbf{Parameter} & \textbf{Value} & \textbf{Parameter} & \textbf{Value} \\
\midrule
\textit{Pretraining epochs} & 10 & \textit{Batch size} & 48 \\
\textit{Learning rate} & $1\times10^{-4}$ & \textit{Weight decay} & $1\times10^{-2}$ \\
\textit{Max length} & 1024 & \textit{Transformer layers} & 6 \\
\textit{Attention heads} & 8 & \textit{Dropout rate} & 0.1 \\
\textit{Box size} & 10 & \textit{Latent dimension} & 128 \\
\textit{Mask ratio} & 0.15 & \textit{$\lambda_1$} & 1.0 \\
\textit{$\lambda_2$} & 0.5 & \textit{$\lambda_3$} & 0.3 \\
\bottomrule
\end{tabular}
\end{table}

\subsection{Fine-tuning on Downstream Tasks}
\label{appendix:fine-tuning}

After pretraining, CryoProt takes the protein sequence as input and infers residue-level map representations through the cross-modal prediction module, eliminating the requirement for experimental cryo-EM density maps during downstream inference. Rather than directly utilizing sequence representations, we exclusively employ the sequence-inferred map representations for downstream tasks, thereby leveraging the density-related structural information learned during multimodal pretraining. For residue-level tasks, a lightweight task-specific prediction head is applied independently to each inferred map representation, preserving fine-grained residue-level information. For protein-level tasks, masked mean pooling is performed over valid residue positions to obtain a global protein representation while excluding padded positions. The resulting representation is subsequently fed into a task-specific prediction head. This strategy enables CryoProt to leverage the knowledge acquired from experimental cryo-EM density maps during pretraining while maintaining sequence-only applicability across diverse downstream tasks.

\subsection{Introduction of Baseline Models}
\label{appendix:baselines}

\subsubsection{Domain-specific Models}

\textbf{Protein Flexibility Prediction}

\begin{itemize}[leftmargin=*, itemsep=0pt, topsep=0pt, parsep=0pt]

\item \textbf{DEFMap} \citep{matsumoto2021extraction} leverages cryo-EM density maps to predict residue-level protein flexibility. It extracts local 3D regions centered at residues from density maps and employs 3D convolutional networks to learn structural patterns associated with flexibility.

\item \textbf{RMSF-net} \citep{song2024accurate} enhances flexibility prediction by combining real cryo-EM density maps with simulated density maps derived from protein structures. It further designs a tailored 3D CNN architecture to capture multi-scale structural features for improved prediction performance.

\item \textbf{ResaPred} \citep{wang2025resapred} is a deep learning-based method that predicts protein flexibility by integrating diverse sequence-derived features, including secondary structure, torsion angles, and solvent accessibility. It adopts a residual network architecture combined with self-attention mechanisms to effectively capture key patterns related to flexibility.

\item \textbf{Flexpert-3D} \citep{kouba2025learning} is a protein flexibility prediction model that incorporates both sequential and structural information. It addresses data limitations and improves performance by capturing flexibility-related patterns from multiple protein feature sources.

\end{itemize}

\textbf{Active Site Identification}

\begin{itemize}[leftmargin=*, itemsep=0pt, topsep=0pt, parsep=0pt]

\item \textbf{Deep-ProBind} \citep{khan2025deep} is a hybrid learning model for identifying protein binding sites by combining sequence-derived and evolutionary features with structural information. It integrates transformer-based encoding with handcrafted descriptors, and employs feature selection together with a neural classifier to perform residue-level binding site prediction.

\item \textbf{MMSite} \citep{ouyang2024mmsite} is a multimodal framework that enhances protein language models by incorporating biomedical language representations. It leverages complementary information from protein sequences and textual biological knowledge to improve residue-level prediction performance.

\item \textbf{M$^3$Site} \citep{ouyang2025m3site} is a multimodal approach for residue-level active site prediction. It integrates protein sequence embeddings, structural graph representations, and functional textual annotations to jointly model biochemical context and improve multiclass active site identification.

\item \textbf{DeepProSite} \citep{fang2023deepprosite} is a structure-aware framework for residue-level protein binding site prediction. It combines predicted structures with pretrained protein language model representations to capture complementary structural and sequence information for identifying residues involved in protein binding.

\end{itemize}

\textbf{Protein--Protein Binding Affinity Prediction}

\begin{itemize}[leftmargin=*, itemsep=0pt, topsep=0pt, parsep=0pt]

\item \textbf{Bind-ddG} \citep{shan2022deep} adopts an attention-based geometric neural network to model the effects of mutations on protein--protein interactions using 3D complex structures. As the training code is not publicly available, we directly use the released pretrained model for inference in our experiments.

\item \textbf{GearBind} \citep{cai2024pretrainable} represents protein--protein interfaces as all-atom graphs and models interactions through multi-level geometric message passing, enabling effective characterization of complex intermolecular interactions.

\item \textbf{Island} \citep{abbasi2020island} is a sequence-driven approach for binding affinity prediction. It utilizes a variety of features derived from protein sequences and applies regression-based learning to capture relationships associated with binding strength.

\item \textbf{PPIgraphomer} \citep{xie2025ppi} is a graph Transformer-based framework for binding affinity prediction. It incorporates pretrained protein representations and models interaction patterns at binding interfaces through refined graph construction and attention mechanisms.

\end{itemize}

\textbf{$\Delta\Delta G$ Prediction}

\begin{itemize}[leftmargin=*, itemsep=0pt, topsep=0pt, parsep=0pt]

\item \textbf{PPIformer} \citep{bushuiev2023learning} is an SE(3)-equivariant model designed to generalize across diverse protein--binder variants. It captures geometric relationships within protein complexes while incorporating a thermodynamically inspired loss formulation to improve the accuracy of prediction.

\item \textbf{ProBASS} \citep{gurusinghe2025probass} leverages both sequence and structural information through pre-trained protein language models ESM2. It generates embeddings for protein–protein interaction mutants and fine-tunes the model on experimentally measured $\Delta\Delta G$ datasets, demonstrating the effectiveness of combining sequence- and structure-based representations for mutation effect prediction.

\item \textbf{DGCddG} \citep{jiang2023dgcddg} is a graph convolution-based approach for predicting mutation-induced changes in binding affinity. It applies multi-layer graph convolutions to learn contextualized residue representations from protein complex structures, and uses these features to model the impact of mutations.

\item \textbf{MFFN} \citep{zhang2025multi} is a multi-scale feature fusion network for $\Delta\Delta G$ prediction that reduces reliance on handcrafted biological features. It models protein complexes at multiple scales and incorporates attention mechanisms along with feature extraction modules to capture diverse structural patterns relevant to binding affinity changes.

\end{itemize}

\subsubsection{Pretrained Models}

\begin{itemize}[leftmargin=*, itemsep=0pt, topsep=0pt, parsep=0pt]

\item \textbf{ESM2} \citep{lin2022language} employs a deeper and more scalable Transformer architecture, trained on millions of protein sequences to effectively capture long-range dependencies and rich evolutionary patterns. In our experiments, we use ESM2-150M as the primary baseline due to its strong performance and moderate computational cost. In addition, we include smaller variants, ESM2-8M and ESM2-35M, in the ablation study to investigate the impact of model scale on downstream performance.

\item \textbf{OmegaFold} \citep{wu2022high} is a protein structure-aware language model that learns informative structural representations. It employs a memory-efficient self-attention architecture to capture spatial dependencies and model three-dimensional protein structures without relying on MSA, providing useful structure-level features for downstream tasks.

\textbf{ESM3} \citep{hayes2025simulating} is a multimodal generative protein language model that jointly represents protein sequences, structures, and functions through discrete token representations. It employs a generative masked modeling objective to learn dependencies across these modalities, enabling both protein generation and representation learning. In our experiments, we use the ESM3-Open-1.4B variant as a baseline.

\textbf{SaProt} \citep{su2024saprot} introduces a structure-aware vocabulary that combines amino acid tokens with structural tokens derived from Foldseek. Pretrained on approximately 40 million protein sequence-structure pairs, it explicitly incorporates structural information into protein language modeling to learn representations that capture both sequence and structural characteristics. We adopt the SaProt-650M-PDB variant for comparison.

\textbf{ProstT5} \citep{heinzinger2024bilingual} is a bilingual protein language model built upon the T5 encoder-decoder architecture. It represents protein structures using Foldseek's 3Di tokens and is trained to perform bidirectional translation between amino acid sequences and structural representations. This approach enables the model to integrate sequence and structural information into transferable protein embeddings.

\textbf{ProtST} \citep{xu2023protst} is a multimodal protein pretraining framework that integrates protein sequences with biomedical text descriptions of protein properties and functions. It combines unimodal masked prediction, multimodal representation alignment, and multimodal masked prediction to incorporate functional knowledge into protein representations while preserving sequence-level information. We use the ProtST-ESM-1b variant in our experiments.

\end{itemize}

\subsection{Evaluation Metrics}
\label{appendix:metrics}

We evaluate CryoProt on four downstream tasks, including regression tasks (e.g., protein flexibility prediction, binding affinity prediction, and $\Delta\Delta G$ prediction) and a binary classification task (active site identification). 

\textbf{Regression Tasks.} We adopt Root Mean Square Error (RMSE), Pearson correlation coefficient (PC), and Spearman rank correlation coefficient (SC). Given ground-truth values $\{y_i\}_{i=1}^{S}$ and predictions $\{\hat{y}_i\}_{i=1}^{S}$, RMSE is defined as
\begin{equation}
\text{RMSE} = \sqrt{\frac{1}{S} \sum_{i=1}^{S} (y_i - \hat{y}_i)^2},
\end{equation}
where $S$ denotes the total number of samples. The Pearson correlation coefficient is computed as
\begin{equation}
\text{PC} = \frac{\sum_{i=1}^{S} (y_i - \bar{y})(\hat{y}_i - \bar{\hat{y}})}{\sqrt{\sum_{i=1}^{S} (y_i - \bar{y})^2} \sqrt{\sum_{i=1}^{S} (\hat{y}_i - \bar{\hat{y}})^2}},
\end{equation}
the Spearman rank correlation coefficient is defined as the Pearson correlation between ranked variables
\begin{equation}
\text{SC} = \frac{\sum_{i=1}^{S} (r_i - \bar{r})(\hat{r}_i - \bar{\hat{r}})}{\sqrt{\sum_{i=1}^{S} (r_i - \bar{r})^2} \sqrt{\sum_{i=1}^{S} (\hat{r}_i - \bar{\hat{r}})^2}},
\end{equation}
where $r_i$ and $\hat{r}_i$ denote the ranks of $y_i$ and $\hat{y}_i$, respectively. For residue-level tasks, $S$ corresponds to the total number of residues across all proteins, whereas for protein-level tasks, $S$ denotes the number of protein samples.

\textbf{Classification Task.} For active site identification, we adopt Area Under the Precision-Recall Curve (AUPRC), Matthews Correlation Coefficient (MCC), and False Positive Rate (FPR). Given binary labels and predictions, we compute true positives (TP), true negatives (TN), false positives (FP), and false negatives (FN), where TP denotes correctly predicted positive samples, TN denotes correctly predicted negative samples, FP denotes incorrectly predicted positive samples, and FN denotes incorrectly predicted negative samples. Based on these quantities, precision and recall are defined as
\begin{equation}
\text{Precision} = \frac{TP}{TP + FP}, \quad
\text{Recall} = \frac{TP}{TP + FN}.
\end{equation}
AUPRC is computed as the area under the precision–recall curve obtained by varying the classification threshold. MCC and FPR are defined as
\begin{equation}
\text{MCC} =
\frac{TP \cdot TN - FP \cdot FN}{
\sqrt{
\begin{gathered}
(TP+FP)(TP+FN) \\
{}\times (TN+FP)(TN+FN)
\end{gathered}
}},
\end{equation}
\begin{equation}
\text{FPR} = \frac{FP}{FP + TN}.
\end{equation}

\begin{table*}[!t]
\centering
\caption{
Ablation studies on protein flexibility prediction and active site identification.
The best results are in bold.
}
\label{appendix:tab:ablation_residue}

\setlength{\tabcolsep}{4pt}
\renewcommand{\arraystretch}{1.0}

\begin{tabular}{c|ccc|c|ccc}
\toprule

\multicolumn{4}{c|}{\textbf{\textit{A. Protein flexibility prediction (residue-level)}}}
&
\multicolumn{4}{c}{\textbf{\textit{B. Active site identification (residue-level)}}} \\

\textbf{Model}
& \textbf{RMSE $\downarrow$}
& \textbf{PC $\uparrow$}
& \textbf{SC $\uparrow$}
& \textbf{Model}
& \textbf{AUPRC $\uparrow$}
& \textbf{MCC $\uparrow$}
& \textbf{FPR $\downarrow$} \\
\midrule

\rowcolor{gray!15}
\multicolumn{4}{c}{\textit{Box interactions}}
&
\multicolumn{4}{c}{\textit{Box interactions}} \\

w/o MLA
& 0.546 & 0.271 & 0.278
& w/o MLA
& 0.640 & 0.575 & 0.026 \\

w/ local-window attention
& 0.519 & 0.291 & 0.296
& w/ local-window attention
& 0.728 & 0.641 & 0.025 \\

w/ self-attention
& 0.497 & 0.302 & 0.307
& w/ self-attention
& 0.738 & 0.651 & 0.023 \\

\rowcolor{gray!15}
\multicolumn{4}{c}{\textit{Cryo-EM maps}}
&
\multicolumn{4}{c}{\textit{Cryo-EM maps}} \\

w/ synthesized maps
& 0.521 & 0.286 & 0.293
& w/ synthesized maps
& 0.724 & 0.636 & 0.025 \\

w/ PDB structure
& 0.536 & 0.276 & 0.284
& w/ PDB structure
& 0.711 & 0.621 & 0.027 \\

w/ sequence
& 0.551 & 0.263 & 0.270
& w/ sequence
& 0.694 & 0.603 & 0.029 \\

\rowcolor{gray!15}
\multicolumn{4}{c}{\textit{Local resolution}}
&
\multicolumn{4}{c}{\textit{Local resolution}} \\

w/ local resolution
& 0.489 & 0.307 & 0.312
& w/ local resolution
& 0.740 & 0.661 & 0.022 \\

\rowcolor{gray!15}
\multicolumn{4}{c}{\textit{Synthesized maps}}
&
\multicolumn{4}{c}{\textit{Synthesized maps}} \\

Synthetic replacement
& 0.521 & 0.286 & 0.293
& Synthetic replacement
& 0.724 & 0.636 & 0.025 \\

Synthetic augmentation
& 0.503 & 0.299 & 0.303
& Synthetic augmentation
& 0.737 & 0.655 & 0.023 \\

\rowcolor{gray!15}
\multicolumn{4}{c}{\textit{Pretraining tasks}}
&
\multicolumn{4}{c}{\textit{Pretraining tasks}} \\

w/o pretraining task 2
& 0.498 & 0.308 & 0.309
& w/o pretraining task 2
& 0.713 & 0.638 & 0.023 \\

w/o pretraining task 3
& 0.508 & 0.302 & 0.306
& w/o pretraining task 3
& 0.701 & 0.624 & 0.023 \\

w/o pretraining tasks 2 + 3
& 0.532 & 0.284 & 0.291
& w/o pretraining tasks 2 + 3
& 0.662 & 0.588 & 0.025 \\

\rowcolor{gray!15}
\multicolumn{4}{c}{\textit{Sequence encoder}}
&
\multicolumn{4}{c}{\textit{Sequence encoder}} \\

w/ ESM-1b
& 0.518 & 0.295 & 0.298
& w/ ESM-1b
& 0.695 & 0.618 & 0.023 \\

w/ ESM2-8M
& 0.509 & 0.301 & 0.304
& w/ ESM2-8M
& 0.688 & 0.611 & 0.024 \\

w/ ESM2-35M
& 0.499 & 0.307 & 0.309
& w/ ESM2-35M
& 0.706 & 0.629 & 0.023 \\

w/ ProtBert
& 0.503 & 0.304 & 0.307
& w/ ProtBert
& 0.718 & 0.641 & 0.022 \\

w/ ProtAlbert
& 0.506 & 0.300 & 0.303
& w/ ProtAlbert
& 0.711 & 0.633 & 0.022 \\

w/ ProtXLNet
& 0.504 & 0.302 & 0.305
& w/ ProtXLNet
& 0.714 & 0.636 & 0.023 \\

\rowcolor{gray!15}
\multicolumn{4}{c}{\textit{Capacity control}}
&
\multicolumn{4}{c}{\textit{Capacity control}} \\

ESM2 + Matched Head
& 0.612 & 0.264 & 0.227
& ESM2 + Matched Head
& 0.695 & 0.628 & 0.038 \\

\midrule
\rowcolor[HTML]{F2F2FF}
\textbf{CryoProt}
& \textbf{0.484}
& \textbf{0.310}
& \textbf{0.315}
& \textbf{CryoProt}
& \textbf{0.746}
& \textbf{0.669}
& \textbf{0.021} \\

\bottomrule
\end{tabular}
\end{table*}

\section{Additional Ablation Studies}
\label{appendix:ablation study}

The effects of cross-box interactions and experimental cryo-EM density maps on protein flexibility prediction are analyzed in Section~\ref{ablation study}. To provide a more comprehensive evaluation of the individual components of CryoProt, we present the full ablation results for this task in Table~\ref{appendix:tab:ablation_residue}. In addition to cross-box interactions and cryo-EM maps, we further investigate the effects of local resolution information, synthesized maps, different pretraining tasks, and alternative sequence encoders. We also include a capacity-matched baseline to examine whether the performance gains can be attributed solely to increased model capacity. These additional experiments provide a more comprehensive understanding of the contributions of individual architectural components and pretraining strategies.

\textbf{Cryo-EM maps.}
We replace experimental cryo-EM maps with ChimeraX-synthesized maps \citep{meng2023ucsf}, PDB structures, or sequence representations while retaining the MLA architecture and pretraining objectives. The PDB-based variant uses ESM-IF1 \citep{hsu2022learning} residue embeddings, whereas the sequence-only variant uses ESM2 embeddings. These replacements increase RMSE from 0.484 to 0.521, 0.536, and 0.551, respectively. The consistent decline across all three metrics supports the contribution of experimental density information.

\textbf{Local resolution.} We calculated voxel-wise local resolution using ResMap\footnote{\url{https://resmap.sourceforge.net/}} and added the resolution value of each voxel as an additional feature channel to the original box representation. Incorporating explicit local resolution information brings only marginal changes, suggesting that CryoProt can already capture local density quality variations from the original density representations.

\textbf{Synthesized maps.}
We examine whether synthesized density maps can replace experimental
maps or expand the pretraining dataset. For synthetic replacement,
we replace all experimental maps with ChimeraX-synthesized maps.
For synthetic augmentation, we add 8,927 synthesized maps for proteins
previously excluded because of their lower map resolution
(4--8~\AA), expanding the pretraining set to 29,457 proteins.
Replacement and augmentation yield RMSE values of 0.521 and 0.503,
respectively, compared with 0.484 for CryoProt. Neither setting
improves performance over pretraining with experimental maps alone.

\textbf{Sequence encoder variants.}
We replace the default ESM2-150M sequence encoder with ESM-1b \citep{rives2021biological}, ESM2-8M, ESM2-35M \citep{lin2022language}, ProtBert, ProtAlbert, or ProtXLNet \citep{elnaggar2021prottrans} to assess sensitivity to encoder choice. ESM2-150M achieves the lowest RMSE and highest correlations among the tested encoders. Within the ESM2 family, RMSE decreases from 0.509 to 0.499 and 0.484 as model size increases from 8M to 35M and 150M parameters, respectively. These results
support the choice of ESM2-150M as the sequence encoder for CryoProt.

\textbf{Capacity control.}
To examine whether the performance gains arise solely from increased model capacity, we introduce a capacity-matched baseline consisting of ESM2-150M and an additional prediction head with the same downstream model capacity as CryoProt. Both models are trained and evaluated under the same protocol. CryoProt outperforms this baseline across all three metrics, reducing RMSE from 0.612 to 0.484. These results indicate that the improvements cannot be explained solely by increased downstream model capacity.

\begin{table*}[!t]
\centering
\caption{
Ablation studies on binding affinity prediction and $\Delta\Delta G$ prediction.
The best results are in bold.
}
\label{appendix:tab:ablation_protein}

\setlength{\tabcolsep}{4pt}
\renewcommand{\arraystretch}{1.0}

\begin{tabular}{c|ccc|c|ccc}
\toprule

\multicolumn{4}{c|}{\textbf{\textit{A. Binding affinity prediction (protein-level)}}}
&
\multicolumn{4}{c}{\textbf{\textit{B. $\Delta\Delta G$ prediction (protein-level)}}} \\

\textbf{Model}
& \textbf{RMSE $\downarrow$}
& \textbf{PC $\uparrow$}
& \textbf{SC $\uparrow$}
& \textbf{Model}
& \textbf{RMSE $\downarrow$}
& \textbf{PC $\uparrow$}
& \textbf{SC $\uparrow$} \\
\midrule

\rowcolor{gray!15}
\multicolumn{4}{c}{\textit{Box interactions}}
&
\multicolumn{4}{c}{\textit{Box interactions}} \\

w/o MLA
& 0.575 & 0.445 & 0.448
& w/o MLA
& 1.295 & 0.548 & 0.438 \\

w/ local-window attention
& 0.528 & 0.478 & 0.486
& w/ local-window attention
& 1.151 & 0.586 & 0.454 \\

w/ self-attention
& 0.507 & 0.503 & 0.509
& w/ self-attention
& 1.096 & 0.602 & 0.476 \\

\rowcolor{gray!15}
\multicolumn{4}{c}{\textit{Cryo-EM maps}}
&
\multicolumn{4}{c}{\textit{Cryo-EM maps}} \\

w/ synthesized maps
& 0.531 & 0.471 & 0.478
& w/ synthesized maps
& 1.164 & 0.579 & 0.448 \\

w/ PDB structure
& 0.548 & 0.451 & 0.459
& w/ PDB structure
& 1.213 & 0.556 & 0.421 \\

w/ sequence
& 0.572 & 0.417 & 0.425
& w/ sequence
& 1.286 & 0.523 & 0.392 \\

\rowcolor{gray!15}
\multicolumn{4}{c}{\textit{Local resolution}}
&
\multicolumn{4}{c}{\textit{Local resolution}} \\

w/ local resolution
& 0.495 & 0.516 & 0.520
& w/ local resolution
& 1.061 & 0.605 & 0.482 \\

\rowcolor{gray!15}
\multicolumn{4}{c}{\textit{Synthesized maps}}
&
\multicolumn{4}{c}{\textit{Synthesized maps}} \\

Synthetic replacement
& 0.531 & 0.471 & 0.478
& Synthetic replacement
& 1.164 & 0.579 & 0.448 \\

Synthetic augmentation
& 0.501 & 0.511 & 0.515
& Synthetic augmentation
& 1.075 & 0.601 & 0.479 \\

\rowcolor{gray!15}
\multicolumn{4}{c}{\textit{Pretraining tasks}}
&
\multicolumn{4}{c}{\textit{Pretraining tasks}} \\

w/o pretraining task 2
& 0.509 & 0.507 & 0.511
& w/o pretraining task 2
& 1.108 & 0.596 & 0.476 \\

w/o pretraining task 3
& 0.528 & 0.492 & 0.498
& w/o pretraining task 3
& 1.162 & 0.585 & 0.468 \\

w/o pretraining tasks 2 + 3
& 0.563 & 0.461 & 0.465
& w/o pretraining tasks 2 + 3
& 1.248 & 0.561 & 0.451 \\

\rowcolor{gray!15}
\multicolumn{4}{c}{\textit{Sequence encoder}}
&
\multicolumn{4}{c}{\textit{Sequence encoder}} \\

w/ ESM-1b
& 0.541 & 0.472 & 0.476
& w/ ESM-1b
& 1.205 & 0.568 & 0.456 \\

w/ ESM2-8M
& 0.532 & 0.483 & 0.487
& w/ ESM2-8M
& 1.168 & 0.579 & 0.463 \\

w/ ESM2-35M
& 0.518 & 0.497 & 0.501
& w/ ESM2-35M
& 1.120 & 0.592 & 0.472 \\

w/ ProtBert
& 0.523 & 0.492 & 0.496
& w/ ProtBert
& 1.135 & 0.586 & 0.468 \\

w/ ProtAlbert
& 0.527 & 0.488 & 0.492
& w/ ProtAlbert
& 1.148 & 0.582 & 0.465 \\

w/ ProtXLNet
& 0.525 & 0.490 & 0.494
& w/ ProtXLNet
& 1.142 & 0.584 & 0.467 \\

\rowcolor{gray!15}
\multicolumn{4}{c}{\textit{Capacity control}}
&
\multicolumn{4}{c}{\textit{Capacity control}} \\

ESM2 + Matched Head
& 0.528 & 0.434 & 0.503
& ESM2 + Matched Head
& 1.289 & 0.554 & 0.411 \\

\midrule
\rowcolor[HTML]{F2F2FF}
\textbf{CryoProt}
& \textbf{0.486}
& \textbf{0.524}
& \textbf{0.527}
& \textbf{CryoProt}
& \textbf{1.036}
& \textbf{0.611}
& \textbf{0.488} \\

\bottomrule
\end{tabular}
\end{table*}

To further evaluate the effectiveness of CryoProt across diverse downstream tasks, we conduct additional ablation studies on active site identification, binding affinity prediction, and $\Delta\Delta G$ prediction, with the detailed results reported in Table~\ref{appendix:tab:ablation_residue} and Table~\ref{appendix:tab:ablation_protein}. Following the same experimental settings as protein flexibility prediction, we systematically evaluate the contributions of different model components and pretraining strategies, including cross-box interactions, experimental cryo-EM density maps, local resolution information, synthesized maps, pretraining tasks, and sequence encoder variants. In addition, capacity control experiments are conducted to distinguish the benefits of the proposed pretraining framework from those arising solely from increased model capacity. The results show that the complete CryoProt framework consistently achieves superior performance over its ablated variants across all three downstream tasks, highlighting the importance of integrating experimental density information and modeling cross-box interactions during pretraining. These additional experiments provide further evidence for the effectiveness of the proposed framework and demonstrate its applicability across both residue-level and protein-level prediction tasks with diverse biological objectives.

\begin{figure*}[!h]
\centering
\includegraphics[width=\linewidth]{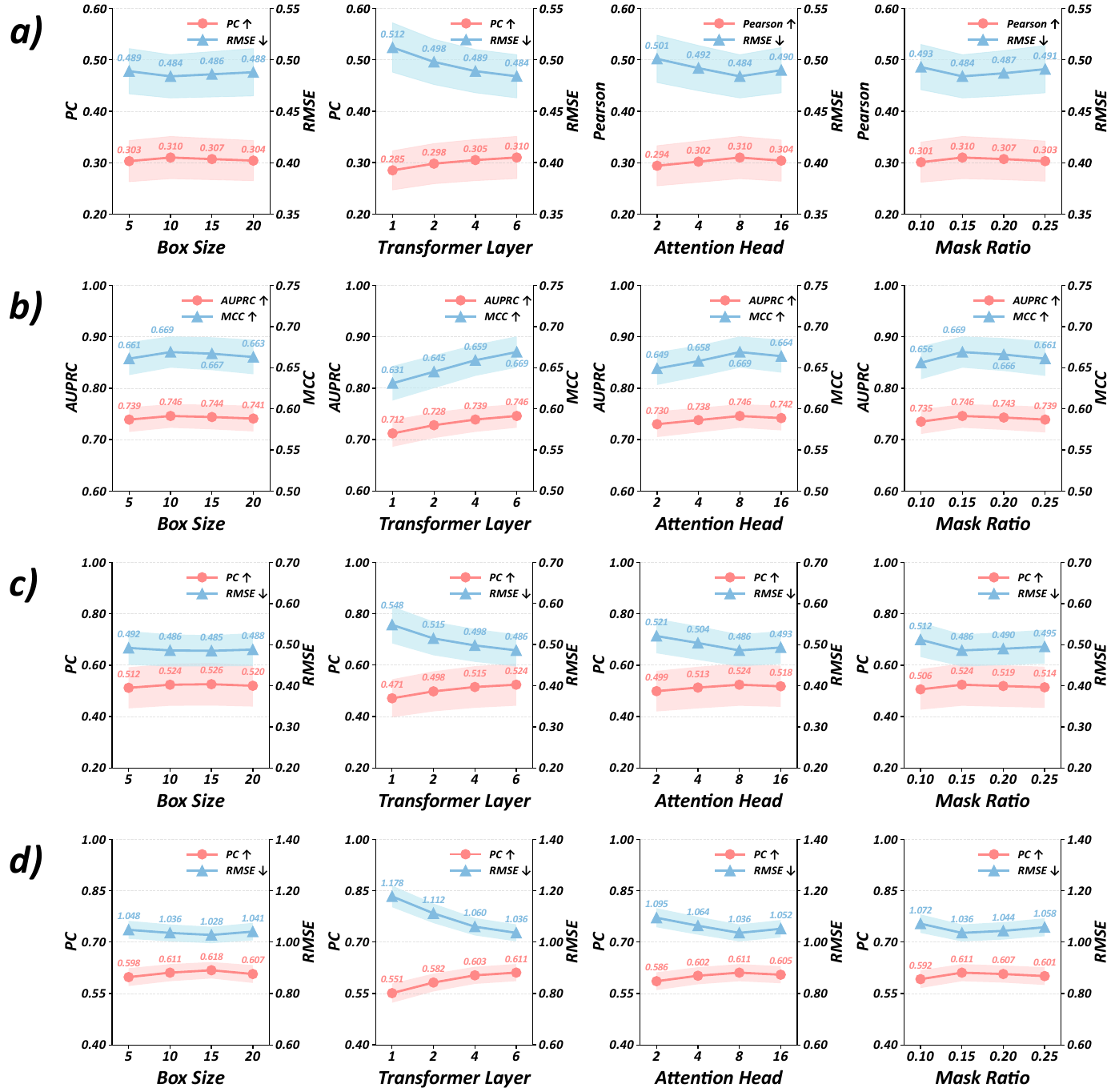}
\caption{Parameter sensitivity analysis of CryoProt with respect to 
four key hyperparameters, including box size, number of Transformer 
layers, number of attention heads, and mask ratio. These 
hyperparameters are associated with the spatial granularity of 
density map partitioning, the depth of cross-box interaction 
modeling, the diversity of attention patterns, and the difficulty 
of masked box reconstruction, respectively. The sensitivity of 
CryoProt to different hyperparameter settings is evaluated across 
four downstream tasks: (a) protein flexibility prediction, 
(b) active site identification, (c) binding affinity prediction, 
and (d) $\Delta\Delta G$ prediction. }
\label{fig:appendix_parameter_sensitivity}
\end{figure*}

\section{Additional Parameter Sensitivity Analysis}
\label{appendix:parameter_sensitivity}

We further investigate the sensitivity of CryoProt to two important 
hyperparameters, namely the number of attention heads and the mask 
ratio, on protein flexibility prediction. These two hyperparameters 
govern the modeling of cross-box interactions and the difficulty of 
masked box reconstruction during pretraining, respectively. The 
experimental results are presented in 
Fig.~\ref{fig:appendix_parameter_sensitivity}.

\textbf{Attention Heads.} Increasing the number of attention 
heads from 2 to 8 consistently improves prediction performance, 
indicating that a moderate number of attention heads enables CryoProt 
to capture diverse spatial dependencies and complementary interaction 
patterns among boxes. With fewer attention heads, the model may have 
limited capacity to simultaneously characterize different types of 
spatial interactions, potentially restricting the learning of complex 
structural relationships. However, when the number of attention heads 
is further increased to 16, the prediction performance slightly 
decreases. This phenomenon suggests that excessive attention heads 
may introduce redundant or overly fragmented interaction patterns 
without providing additional useful information. Moreover, under a 
fixed embedding dimension, increasing the number of heads reduces 
the feature dimension of each head, potentially limiting its 
representational capacity. Based on these observations, we adopt 
8 attention heads as the default configuration to achieve a balance 
between interaction diversity and representation quality.

\textbf{Mask Ratio.} CryoProt achieves the best overall performance 
when the mask ratio is set to 0.15, highlighting the importance of 
selecting an appropriate reconstruction difficulty during pretraining. 
A smaller mask ratio leaves most spatial boxes visible, which may 
make the reconstruction task relatively straightforward and reduce 
the necessity of learning informative contextual dependencies. 
Conversely, a larger mask ratio removes more local density information, 
potentially disrupting spatial continuity and limiting the contextual 
information available for reconstructing masked boxes. This may 
increase the difficulty of learning meaningful density-related 
structural representations. A moderate mask ratio encourages the 
model to exploit contextual information and cross-box dependencies 
while preserving sufficient spatial information for effective 
reconstruction. Therefore, we adopt a mask ratio of 0.15 as the 
default setting, providing a favorable balance between reconstruction 
difficulty and contextual information availability.

\section{Consistency Between Inferred and Real Map Representations}
\label{appendix:map_consistency}

To examine whether sequence-inferred map representations retain information encoded from experimental cryo-EM maps, we compare them with representations extracted directly by the Map Encoder. We use the same 120 proteins as in the Top-$L$ coverage study, with 40 proteins in each of the 2--4, 4--6, and 6--8~\AA\ resolution ranges. Cosine similarity measures directional alignment in the representation space, while Pearson correlation assesses correspondence in feature variation after mean centering. Agreement under both metrics therefore characterizes shared organization between the inferred and map-derived representations.

\begin{figure}[!h]
\centering
\includegraphics[width=\linewidth]{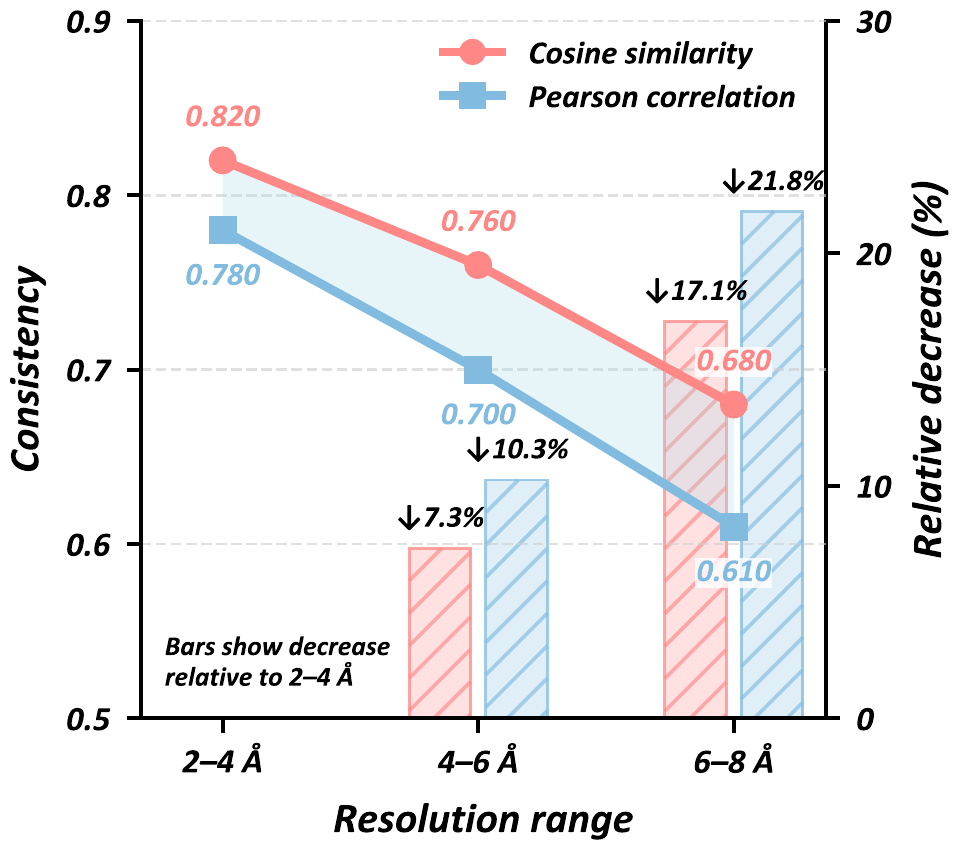}
\caption{Consistency between sequence-inferred and real map representations.}
\label{fig:map_consistency}
\end{figure}

As shown in Fig.~\ref{fig:map_consistency}, the average cosine similarity and Pearson correlation are 0.82 and 0.78 in the 2--4~\AA\ range. These values decrease to 0.76 and 0.70 at 4--6~\AA, and to 0.68 and 0.61 at 6--8~\AA. Relative to the highest-resolution group, the intermediate-resolution group exhibits decreases of 7.3\% and 10.3\%, respectively. The corresponding decreases reach 17.1\% and 21.8\% in the lowest-resolution group, equivalent to absolute reductions of 0.14 and 0.17. Pearson correlation shows a larger relative decline across both comparisons, indicating that the observed resolution dependence is stronger under this metric.

The decline in both metrics is consistent with reduced structural detail in lower-resolution maps, which may make map-derived representations harder to infer from sequence. The comparison across resolution groups establishes an association, with protein composition and map quality also potentially contributing to the observed differences. Positive alignment remains in the 6--8~\AA\ group, suggesting that the sequence-based predictor retains correspondence with map-derived features even in this resolution range. These results support the intended transfer of map-related information into sequence-inferred representations and show how this correspondence varies with map resolution, while downstream inference proceeds from sequence alone.

\begin{table*}[!h]
\centering
\caption{Top-$L$ Coverage across proteins from different resolution ranges.}
\renewcommand{\arraystretch}{1.0}
\label{tab:per-protein TopL}
\normalsize
\begin{tabular}{c | l l l l l}
\toprule
\textbf{Rsolution range} & \multicolumn{5}{c}{\textbf{Proteins (PDB ID / Top-$L$ Coverage)}} \\
\midrule

\multirow{8}{*}{2--4 Å}
& \texttt{9CZL/0.265} & \texttt{7SK9/0.289} & \texttt{5V7V/0.145} & \texttt{7DQA/0.089} & \texttt{9CB9/0.185} \\
& \texttt{8J8H/0.072} & \texttt{8JPD/0.196} & \texttt{8XQI/0.104} & \texttt{8X63/0.274} & \texttt{7SNF/0.041} \\
& \texttt{8FY4/0.082} & \texttt{9KCP/0.506} & \texttt{9DMY/0.416} & \texttt{9QVF/0.008} & \texttt{8Y0Q/0.106} \\
& \texttt{8EHS/0.000} & \texttt{9U9B/0.144} & \texttt{6Z8D/0.017} & \texttt{9JA5/0.026} & \texttt{9HIX/0.264} \\
& \texttt{7WL3/0.151} & \texttt{7NSL/0.454} & \texttt{9PKV/0.186} & \texttt{9JBO/0.647} & \texttt{9T3P/0.000} \\
& \texttt{8B3A/0.000} & \texttt{9J97/0.057} & \texttt{7STF/0.250} & \texttt{7P1K/0.072} & \texttt{9UT9/0.121} \\
& \texttt{9MOQ/0.462} & \texttt{8XCY/0.298} & \texttt{6PUZ/0.057} & \texttt{8SFJ/0.064} & \texttt{9NOU/0.132} \\
& \texttt{7XBX/0.164} & \texttt{3J89/0.000} & \texttt{8G4E/0.068} & \texttt{9J12/0.058} & \texttt{6WUJ/0.160} \\

\midrule

\multirow{8}{*}{4--6 Å}
& \texttt{8E4C/0.113} & \texttt{5LVC/0.107} & \texttt{7B6Y/0.375} & \texttt{8TIN/0.323} & \texttt{8VID/0.151} \\
& \texttt{6XE6/0.200} & \texttt{8W5Q/0.090} & \texttt{9VKS/0.135} & \texttt{8T53/0.067} & \texttt{7RD8/0.280} \\
& \texttt{8Y32/0.071} & \texttt{9IMH/0.000} & \texttt{7YOZ/0.005} & \texttt{7JLX/0.002} & \texttt{7WRJ/0.065} \\
& \texttt{6EGX/0.124} & \texttt{9R5K/0.057} & \texttt{6R1T/0.048} & \texttt{8I8C/0.025} & \texttt{3JBF/0.087} \\
& \texttt{5H1C/0.008} & \texttt{7OZR/0.072} & \texttt{7NIQ/0.250} & \texttt{6T6V/0.345} & \texttt{5A6G/0.649} \\
& \texttt{7TFS/0.144} & \texttt{9HM6/0.057} & \texttt{8VE6/0.009} & \texttt{8AY4/0.117} & \texttt{7OZ3/0.009} \\
& \texttt{6OFJ/0.095} & \texttt{8JTS/0.306} & \texttt{6V6C/0.159} & \texttt{7TT7/0.152} & \texttt{9B3H/0.096} \\
& \texttt{7RO5/0.468} & \texttt{7SSG/0.175} & \texttt{6MI8/0.203} & \texttt{8PQ5/0.132} & \texttt{9FH4/0.588} \\

\midrule

\multirow{8}{*}{6--8 Å}
& \texttt{8QI0/0.276} & \texttt{5NL2/0.066} & \texttt{5ABB/0.339} & \texttt{3J15/0.162} & \texttt{7QE0/0.118} \\
& \texttt{7PGW/0.046} & \texttt{7PEZ/0.044} & \texttt{6WJG/0.058} & \texttt{9NTT/0.187} & \texttt{6RRT/0.012} \\
& \texttt{8QHV/0.307} & \texttt{8C9C/0.344} & \texttt{3JBJ/0.052} & \texttt{5FL2/0.085} & \texttt{9EAD/0.119} \\
& \texttt{6LM3/0.195} & \texttt{9DVC/0.090} & \texttt{3J5V/0.177} & \texttt{6BU9/0.168} & \texttt{8I2H/0.271} \\
& \texttt{8PHT/0.000} & \texttt{6HWW/0.006} & \texttt{7FIF/0.217} & \texttt{8FG2/0.049} & \texttt{8WJO/0.058} \\
& \texttt{8I21/0.054} & \texttt{8ETT/0.111} & \texttt{4CG6/0.295} & \texttt{7XXL/0.076} & \texttt{9TQB/0.032} \\
& \texttt{6N89/0.099} & \texttt{8E3C/0.137} & \texttt{7YLM/0.052} & \texttt{8Z9V/0.141} & \texttt{8SL3/0.174} \\
& \texttt{8FMB/0.141} & \texttt{3MFP/0.283} & \texttt{2I68/0.013} & \texttt{9FTM/0.000} & \texttt{8BGG/0.072} \\

\bottomrule
\end{tabular}
\end{table*}

\section{Case on Top-$L$ Coverage}
\label{appendix:case_on_topL}

\textbf{Definition of Top-$L$ Coverage.}
Given residue embeddings $\{\mathbf{E}_i\}_{i=1}^{L}$, where $L$ is the protein sequence length, we compute the pairwise cosine similarity:
\begin{equation}
S_{ij} =
\frac{\mathbf{E}_i \cdot \mathbf{E}_j}
{\|\mathbf{E}_i\|_2 \|\mathbf{E}_j\|_2}.
\end{equation}
We select the top-$L$ residue pairs with the highest similarity scores from the strict upper triangle ($1 \leq i < j \leq L$), excluding self-pairs and duplicate pairs, and denote this set as $\mathcal{P}_{\text{top}}$.
From the native structure, we compute the residue distance matrix $\mathbf{D}$, where $D_{ij}$ is the C$_\alpha$ distance between residues $i$ and $j$.
Top-$L$ coverage is defined as
\begin{equation}
\text{Top-}L =
\frac{1}{L}
\sum_{(i,j) \in \mathcal{P}_{\text{top}}}
\mathbb{I}(D_{ij} < 10\,\text{\AA}),
\end{equation}
where $\mathbb{I}(\cdot)$ is the indicator function.
A higher Top-$L$ coverage indicates that a larger fraction of the selected residue pairs are spatially proximal.

\textbf{Resolution setting.}
The resolution ranges used for pretraining and the Top-$L$ analysis serve different purposes. During pretraining, we restrict cryo-EM maps to the 2--4~\AA\ range to ensure sufficiently reliable local density information for box construction and representation learning. In contrast, the Top-$L$ analysis is designed as an additional generalization study to examine whether the learned representations can preserve residue-level spatial relationships for proteins associated with a broader range of cryo-EM resolutions. We therefore evaluate proteins from 2--4~\AA, 4--6~\AA, and 6--8~\AA\ intervals. Importantly, proteins from the lower-resolution intervals are used only for evaluation and are not introduced as additional pretraining data.

\textbf{Interpretation of Top-$L$ Coverage.}
Top-$L$ coverage is used as an unsupervised probe of whether highly similar residue embeddings correspond to spatially proximal residue pairs, rather than as a direct measure of structure reconstruction. Since only a small fraction of all residue pairs are within 10~Å, relatively low absolute coverage values are expected without explicit distance or contact supervision. Nevertheless, CryoProt consistently outperforms sequence-only ESM2 representations, achieving a 16.2\% relative improvement over the ESM2. This indicates that cryo-EM-aware pretraining enables the learned representations to capture richer residue-level spatial relationships.

\textbf{Per-protein Top-$L$ Coverage.}
To provide a more detailed case analysis, we report per-protein Top-$L$ coverage results across different resolution ranges. Specifically, we randomly select 40 proteins from each resolution interval (2--4~\AA, 4--6~\AA, and 6--8~\AA), resulting in a total of 120 proteins. For each protein, we report its PDB ID together with the corresponding Top-$L$ coverage obtained by CryoProt, as summarized in Table~\ref{tab:per-protein TopL}. 

\textbf{Visualization of Residue Distances and Embedding Similarities.}
The corresponding visualizations are presented in Fig.~\ref{fig:case_on_rmsf} in Section~\ref{case}. In the distance maps, red regions indicate spatially proximal residue pairs, whereas blue regions indicate larger separations. In the similarity maps, blue regions represent higher embedding similarity. If the embeddings capture spatial relationships, red regions in the distance maps should therefore tend to correspond to blue regions in the similarity maps. CryoProt exhibits structured similarity patterns beyond the main diagonal, suggesting that its representations encode relationships between distinct residues. The accompanying Top-$L$ coverage provides a quantitative assessment of how well highly similar embeddings identify spatially proximal pairs. CryoProt achieves higher coverage in all three examples, with a smaller advantage in the lowest-resolution case. These results support greater enrichment of spatially proximal residue pairs among the most similar embeddings following cryo-EM-aware pretraining.

\onecolumn

\section{More Visualization Results}
\label{appendix:more_visualization}

\begingroup
\centering

\includegraphics[
    width=\textwidth,
    height=0.78\textheight,
    keepaspectratio
]{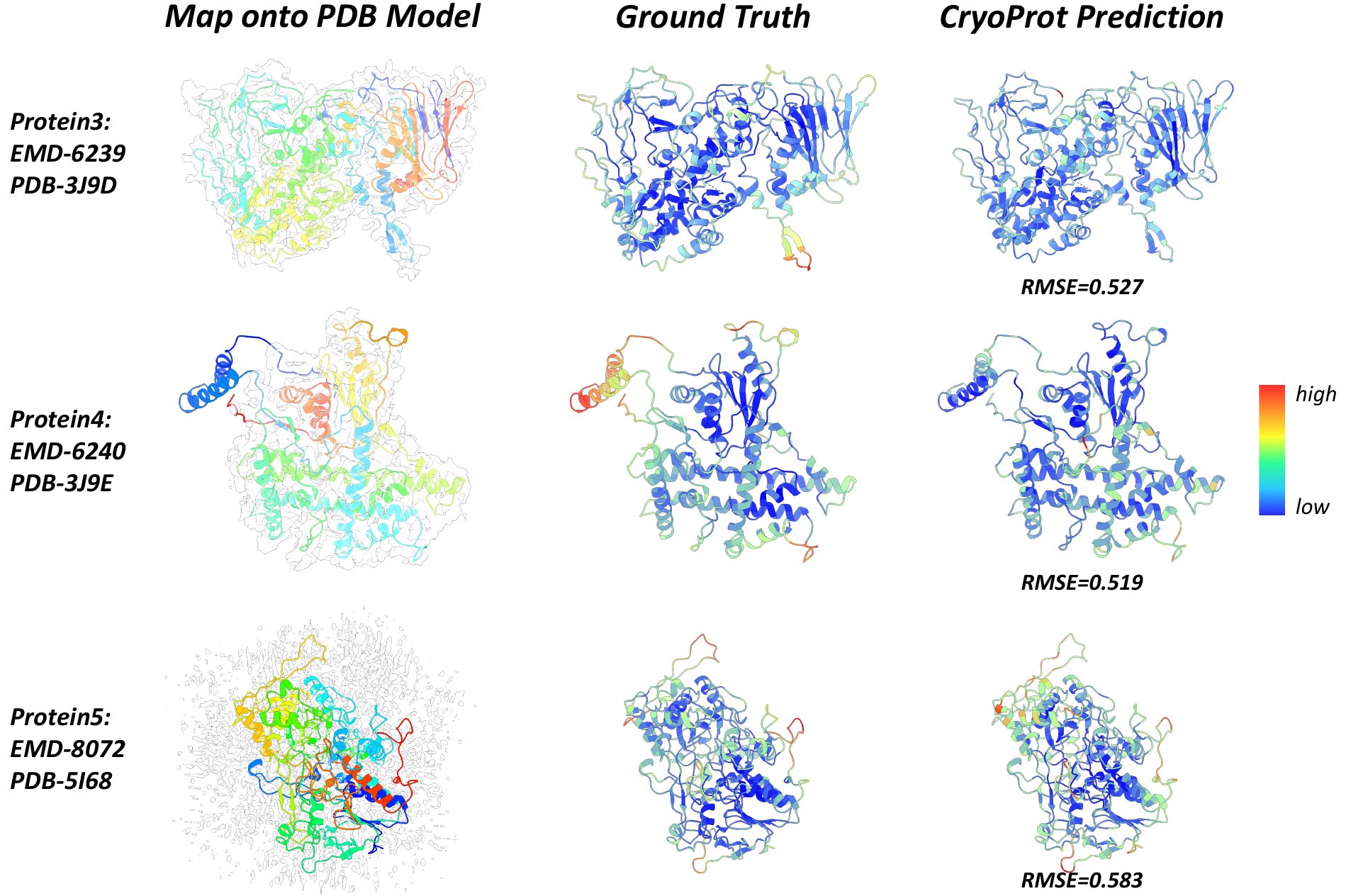}

\captionof{figure}{
    Additional visualization results for protein flexibility prediction.
}
\label{fig:appendix_case_on_rmsf}
\par
\endgroup

\clearpage


\end{document}